\documentclass[10pt,twocolumn,letterpaper]{article}

\usepackage{iccv}              % To produce the CAMERA-READY version
\usepackage{bm}
\usepackage{graphicx}
\definecolor{iccvblue}{rgb}{0.21,0.49,0.74}
\usepackage[pagebackref,breaklinks,colorlinks,allcolors=iccvblue]{hyperref}
\usepackage{endnotes}
\usepackage{multirow}
\def\paperID{4492} % *** Enter the Paper ID here
\def\confName{ICCV}
\def\confYear{2025}

\title{UniFuse: A Unified All-in-One Framework for Multi-Modal Medical Image Fusion Under Diverse Degradations and Misalignments}

\author{%
  Dayong Su\textsuperscript{1}, %
  Yafei Zhang\textsuperscript{1}, %
  Huafeng Li\textsuperscript{1}\thanks{Corresponding author.}, %
  Jinxing Li\textsuperscript{2}, %
  Yu Liu\textsuperscript{3}%
  \\
  \textsuperscript{1}Kunming University of Science and Technology,\\
  \textsuperscript{2}Harbin Institute of Technology at Shenzhen,
  \textsuperscript{3}Hefei University of Technology\\
  {\tt\small dayongsu@outlook.com, \{zyfeimail, hfchina99\}@163.com, }\\
  {\tt\small lijinxing158@hit.edu.cn,  yuliu@hfut.edu.cn}\\
}

\begin{document}
\maketitle
\begin{abstract}
Current multimodal medical image fusion typically assumes that source images are of high quality and perfectly aligned at the pixel level. Its effectiveness heavily relies on these conditions and often deteriorates when handling misaligned or degraded medical images. To address this, we propose UniFuse, a general fusion framework. By embedding a degradation-aware prompt learning module, UniFuse seamlessly integrates multi-directional information from input images and correlates cross-modal alignment with restoration, enabling joint optimization of both tasks within a unified framework. Additionally, we design an Omni Unified Feature Representation scheme, which leverages Spatial Mamba to encode multi-directional features and mitigate modality differences in feature alignment. To enable simultaneous restoration and fusion within an All-in-One configuration, we propose a Universal Feature Restoration \& Fusion module, incorporating the Adaptive LoRA Synergistic Network (ALSN) based on LoRA principles. By leveraging ALSN’s adaptive feature representation along with degradation-type guidance, we enable joint restoration and fusion within a single-stage framework. Compared to staged approaches, UniFuse unifies alignment, restoration, and fusion within a single framework. Experimental results across multiple datasets demonstrate the method’s effectiveness and significant advantages over existing approaches.code is available at https://github.com/slrl123/UniFuse.
\end{abstract}\vspace{-4.5mm}    
\section{Introduction}
\label{sec:intro}

%With the rapid development of medical imaging technology, multimodal medical image fusion has become increasingly important in medical diagnosis and treatment. By combining images from different modalities, the strengths of each modality can be leveraged to improve image quality, thereby enabling doctors to make more accurate diagnoses. Despite extensive research efforts to address growing clinical demands, as demonstrated by recent advancements, including AMMNet \cite{1}, F-DARTS \cite{2}, MATR \cite{3}, and FusionMamba \cite{4}, these methods typically assume that source images are of optimal quality and perfectly aligned. However, in real-world medical imaging, image quality can degrade due to factors such as noise, patient motion, or equipment limitations, and spatial misalignment is common, especially during dynamic scanning or multi-device imaging. As a result, high-quality medical images suitable for direct fusion are often unavailable in practice.
With advancements in medical imaging, multimodal image fusion enhances diagnosis and treatment by integrating modality-specific strengths to improve image quality. While methods like AMMNet \cite{1}, F-DARTS \cite{17}, MATR \cite{3}, and FusionMamba \cite{4} have made progress, they assume high-quality, perfectly aligned source images. In reality, image quality often degrades due to noise, motion, or equipment limitations, and spatial misalignment is common in dynamic scanning or multi-device imaging, making direct fusion challenging.
\begin{figure}[t!]
	\centering
	\includegraphics[height=2.2in,width=2.9in]{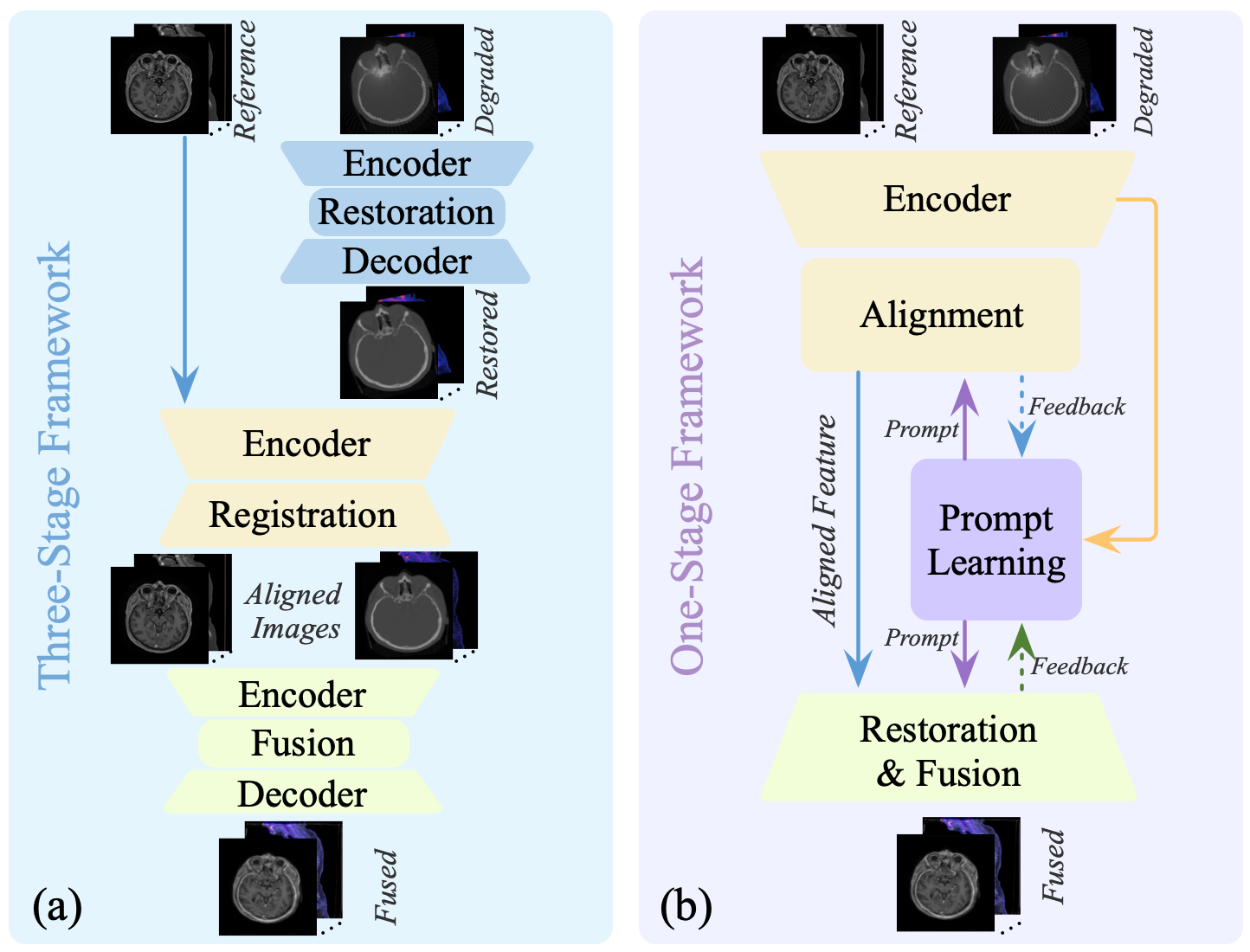}\vspace{-2.5mm}
	\caption{Difference in processing paradigms between the proposed method and existing methods. (a) Processing paradigm of existing methods, (b) Processing paradigm of the proposed method.}\vspace{-4.5mm}
	\label{fig1}
\end{figure}

Traditional solutions to this problem mostly rely on a staged processing framework, typically separating the tasks of restoration, alignment, and fusion. Although this approach can be effective, it often involves combining independent models for each task, as shown in Figure \ref{fig1}(a). These models are usually trained separately, leading to a lack of effective collaboration and mutual reinforcement among the tasks. Consequently, the trained models may not necessarily perform optimally when applied to downstream tasks. Additionally, this paradigm faces challenges such as high computational complexity and difficulties in efficiently managing multiple tasks simultaneously. Therefore, addressing the multimodal medical image fusion, which involves multiple degradations and misalignments, within a unified framework is of significant practical importance.

For the fusion of unaligned and degraded multimodal medical images, we make the first attempt to achieve this within a single-stage framework, as shown in Figure \ref{fig1}(b). Specifically, we propose a general fusion framework named UniFuse. This framework integrates multi-directional information from input images by embedding a degradation-aware prompt learning (DAPL) module, which constructs a shared prompt based on these features to facilitate both cross-modal feature alignment and restoration. With the shared prompt serving as a bridge, the relationship between cross-modal feature alignment and restoration is established, enabling joint optimization of both tasks within a unified framework. Furthermore, by leveraging the multi-directional information integrated by the DAPL, we propose Omni Unified Feature Representation (OUFR) method. This method effectively embeds multi-directional information from source images by introducing Spatial Mamba, addressing challenges arising from modality differences in cross-modal feature alignment.

To simultaneously achieve feature restoration and fusion in an All-in-One setting, we propose a Universal Feature Restoration \& Fusion (UFR\&F) module and embed an Adaptive LoRA Synergistic Network (ALSN), which is based on the LoRA principle. ALSN possesses an adaptive feature representation capability that, combined with degradation-type guidance, enables joint processing of feature restoration and fusion within a single-stage framework. This is accomplished without significantly increasing the model’s parameter size. Compared to traditional staged methods, UniFuse integrates feature alignment, restoration, and fusion into a unified framework, efficiently handling multiple tasks while significantly improving computational efficiency and maintaining a low parameter count. Experimental results across multiple datasets demonstrate the high effectiveness of the proposed method, achieving substantial advantages over existing approaches. In summary, the main contributions of this paper are as follows:
\begin{itemize}
	\item We propose a general multimodal medical image fusion framework, \textit{UniFuse}, that effectively addresses image quality degradation and pixel misalignment.
	\item We design the DAPL module, which establishes a strong connection between feature alignment, restoration, and fusion, enabling collaborative training of these tasks within a unified framework.
	\item We propose the OUFR method, which, with the assistance of DAPL, mitigates modality differences in feature alignment using the Spatial Mamba technique.
	\item We design the UFR\&F module, which leverages the Adaptive LoRA Synergistic Network to enhance the efficiency of feature restoration and fusion while effectively controlling model parameter growth.
\end{itemize}\vspace{-1.5mm}

\section{Related Work}\vspace{-1.5mm}
\subsection{Medical Image Restoration}\vspace{-1.5mm}
In medical image acquisition, various factors can lead to image quality degradation. To address this, several restoration methods have been proposed. For instance, to correct motion artifacts in MRI images, MC-CDic \cite{5} employs convolutional dictionary modeling to restore high-quality features. To mitigate the impact of domain shift between training and testing sets, AdaDiff \cite{6} introduces an adaptive diffusion prior, enhancing model generalization. Furthermore, FedPR \cite{7} integrates federated learning and prompt learning to restore degraded MRI images while preserving privacy. For CT images, which are often affected by noise and metal artifacts, DuDoUFNet \cite{8}, Quad-Net \cite{9}, and PND-Net \cite{10} adopt dual-domain restoration strategies for denoising and artifact removal. These methods extract useful information from the sinogram domain to aid in restoring the image domain, demonstrating strong denoising and artifact suppression capabilities. OSCNet \cite{23} introduces a direction-sharing convolutional representation, effectively modeling the shape of metal artifacts and achieving significant artifact suppression. For PET images, which often suffer from noise due to low-dose contrast agents, SDLJ \cite{11} proposes a joint PET/CT denoising method. By incorporating CT image denoising into the PET denoising process, this method leverages complementary features between the two modalities to enhance denoising performance. To address the high computational cost of diffusion-based PET denoising, CDAG \cite{12} adopts a coarse-to-fine diffusion recovery strategy, significantly reducing computational overhead while maintaining denoising effectiveness. However, these methods are designed to handle a single type of degradation, limiting their generalizability.
\begin{figure*}[t!]
	\centering
	\includegraphics[height=3.15in,width=6.3in]{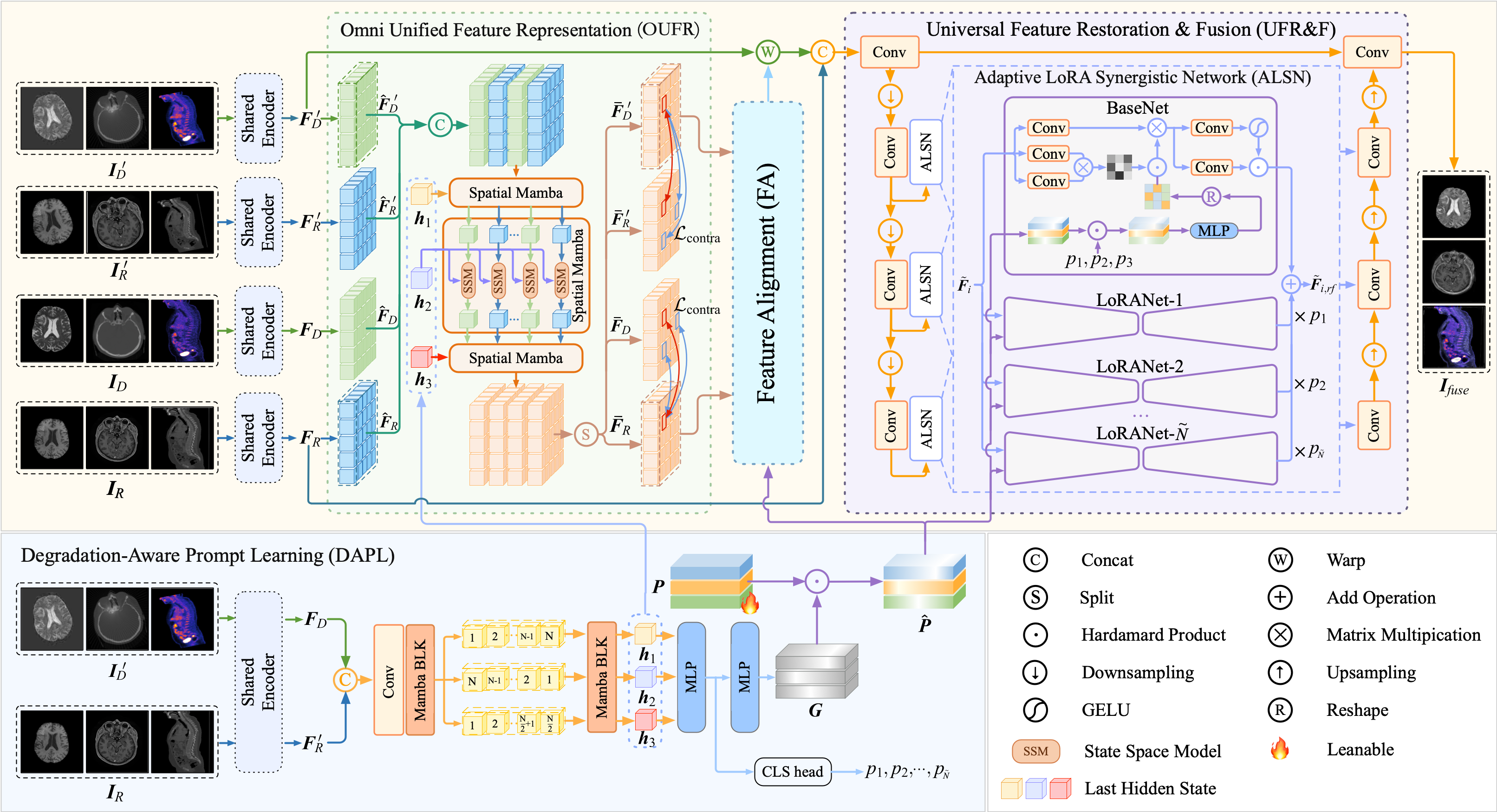}\vspace{-2.5mm}
	\caption{Overview of the proposed framework: The image pair $\left\{ {\bm{I}'_D, \bm{I}_R} \right\}$ is processed by DAPL to obtain multi-directional feature representations ${\bm{h}_1}$, ${\bm{h}_2}$, ${\bm{h}_3}$, and generate a degradation prompt $\bm{\hat{P}}$. In OUFR, the spatial Mamba utilizes the representations to eliminate modal differences between ${\bm{\bar{F}}'_D}$ and ${\bm{\bar{F}}_R}$. FA aligns the source image features. In UFR\&F, an adaptive collaborative network comprising a base network and a LoRA branch uses the degradation cue for the restoration and fusion of various degraded images.}\vspace{-4.5mm}
	\label{fig2}
\end{figure*}

While All-in-One medical image restoration algorithms, such as AMIR \cite{13} and Restore-RWKV \cite{14}, improve generalization, they typically rely on multi-expert ensemble strategies. Although effective, these strategies increase parameter size, reducing model efficiency and deployability. Moreover, these All-in-One methods are primarily designed for 2D images, making it challenging to ensure visual consistency across slices in 3D image volumes. \vspace{-1.5mm}

\subsection{Multimodal Medical Image Fusion}\vspace{-1.5mm}
Multimodal medical image fusion primarily falls into two categories: traditional image fusion methods, which assume pre-registered source images, and joint registration-fusion methods. In traditional fusion methods, CNN-based approaches \cite{1,16,17} leverage the strong representational capabilities of convolutions for local features, thereby enriching fused images with fine details. However, due to the limited receptive field of CNNs, they struggle to capture long-range dependencies effectively. To address this limitation, some methods \cite{3,19,20,21,22} integrate Transformers into medical image fusion networks alongside CNNs. To reduce the computational burden associated with Transformers, FusionMamba \cite{4} replaces them with Mamba in the fusion network, significantly lowering computational costs. While these methods are effective, they assume perfect alignment of source images, which is often unrealistic in practice.

To address spatial misalignment, a staged approach—first registration, then fusion—is commonly used. However, training registration and fusion separately prevents interaction and mutual enhancement, limiting performance gains. Consequently, joint registration-fusion frameworks have gained attention, leading to the development of representative multimodal image fusion methods such as MURF \cite{25}, RECONet \cite{26}, RFVIF \cite{31}, IVFWSR \cite{30}, PAMRFuse \cite{29}, BSAFusion \cite{27}, and MulFS-CAP \cite{28}. However, RFVIF, IVFWSR, and MulFS-CAP are not specifically designed for multimodal medical image fusion and may yield suboptimal results. While MURF, RECONet, BSAFusion, and PAMRFuse address multimodal medical image fusion, they primarily focus on 2D slice data and struggle with spatial shifts in 3D medical imaging. Moreover, these methods assume high-quality source images, making them unsuitable for degraded inputs. To overcome these limitations, this paper proposes UniFuse, a unified framework for fusing unaligned, degraded multimodal medical images. \vspace{-2.5mm}

\section{Methodology}
\label{sec:formatting}
\subsection{Overview}
As shown in Figure \ref{fig2}, the overall framework consists of four core modules: DAPL, OUFR, Feature Alignment (FA), and UFR\&F. DAPL serves two key functions: (1) extracting comprehensive features from input image pairs to reduce modality differences, and (2) identifying degradation types and constructing degradation prompts, enabling the model to adaptively process images based on their degradation. Subsequently, OUFR eliminates modality divergence between source image features, mitigating its negative impact on cross-modal feature alignment. Furthermore, FA further refines this alignment by adaptively predicting spatial shifts through multi-scale modeling, while integrating degradation prototype prompts for precise feature alignment. Lastly, UFR\&F restores and fuses degraded features in a single-stage unified framework, guided by the identified degradation types.

\subsection{Degradation-Aware Prompt Learning}
In DAPL, we employ a Shared Encoder consisting of two convolutional layers to encode the source images $\{ {\bm{I}'_D, \bm{I}_R}\}$. In this process, we assume that the input image $\bm{I}'_D$ is a degraded source image, while the image from the other modality, $\bm{I}_R$, is a high-quality reference image, with a pixel offset between the two. Our method treats $\bm{I}_R$ as the reference image, allowing $\bm{I}'_D$ to be aligned with $\bm{I}_R$ after correction. The outputs of the Shared Encoder for $\bm{I}'_D$ and $\bm{I}_R$ are $\bm{F}'_D \in \mathbb{R}^{C \times H \times W}$ and $\bm{F}_R \in \mathbb{R}^{C \times H \times W}$, respectively, where $C$, $H$, and $W$ denote the number of feature channels, height, and width, respectively. To obtain feature representations that reflect image degradation, we concatenate $\bm{F}'_D$ and $\bm{F}_R$ along the channel dimension and feed the result into a convolutional layer. The output is then patch-encoded and passed through the first Mamba block, yielding $\bm{F}_{for} = [\bm{f}^1, \bm{f}^2, \cdots, \bm{f}^N] \in \mathbb{R}^{N \times M}$, where $N$ denotes the number of patch blocks, and $M$ represents the vector length to which each patch block is mapped.

To obtain feature representations of image degradation from different perspectives, we rearrange the feature $\bm{F}_{for}$ along various directions, resulting in the reverse-ordered feature $\bm{F}_{bac} = [{\bm{f}}^N, {\bm{f}}^{N-1}, \cdots, {\bm{f}}^1]$ and the bidirectional-ordered feature $\bm{F}_{bid} = [{\bm{f}}^1, {\bm{f}}^2, \cdots, {\bm{f}}^{\left\lfloor {N/2} \right\rfloor - 1}, {\bm{f}}^N, \cdots, {\bm{f}}^{\left\lfloor {N/2} \right\rfloor}]$. These three sets of features are then sequentially fed into the second MambaBlock, producing hidden states $\bm{h}_1$, $\bm{h}_2$, and $\bm{h}_3$. Since each hidden state is derived from differently ordered features after further feature extraction, $\bm{h}_1$, $\bm{h}_2$, and $\bm{h}_3$ capture global feature information from different directional perspectives of the input image. It is worth noting that we do not apply the aforementioned reordering to the input feature $\bm{F}_{for}$ in the first MambaBlock. This ensures that $\bm{F}_{bac}$ and $\bm{F}_{bid}$ retain information from $\bm{F}_{for}$, thereby enriching their information content.

To encode degradation information into $\bm{h}_1$, $\bm{h}_2$, and $\bm{h}_3$ so that they can guide cross-modal alignment and feature restoration in subsequent processes, we pass these feature representations through an MLP, followed by a linear layer and a classification head with a Softmax function to obtain the classification result $\bm{p} = [p_1, p_2, \cdots, p_{\tilde{N}}]$, where $\tilde{N}$ represents the number of image degradation categories. Based on $\bm{p}$, we optimize the network parameters using the cross-entropy (CE) loss, defined as
\begin{equation}\small
	{\cal L}_{ce} = \text{CE}(\bm{p}, \bm{p}_{GT}),
\end{equation}
where $\bm{p}_{GT}$ is the ground truth of $\bm p$.
After optimization using Eq. (1), the network outputs $\bm{h}_1$, $\bm{h}_2$, and $\bm{h}_3$, which not only encode degradation information but also provide additional complementary information beneficial for subsequent tasks. Therefore, $\bm{h}_1$, $\bm{h}_2$, and $\bm{h}_3$ can be used not only to guide feature restoration but also to mitigate the impact of modality differences on feature alignment.

Due to the differing requirements for network output when aligning and restoring source images with various degradation types, this paper proposes a degradation-aware auxiliary multi-task joint processing method to reconcile these distinct demands. Specifically, $\bm{h}_1$, $\bm{h}_2$, and $\bm{h}_3$ are processed through two MLPs to generate a prompt selection matrix $\bm{G} \in \mathbb{R}^{\tilde{N} \times L \times M}$. This matrix is then used to select prompts $\bm{\hat{P}}$ from the degradation-aware prompt set $\bm{P} \in \mathbb{R}^{\tilde{N} \times L \times M}$:
\begin{equation}\small
	\bm{\hat{P}} = \bm{P} \odot \bm{G},
\end{equation}
where $\odot$ denotes the Hadamard product. By sharing the same prompt $\bm{\hat{P}}$ for both tasks, this approach establishes a connection between feature cross-modal alignment and feature restoration. It also reduces computational overhead, decreases the parameter scale, and improves inference efficiency. Furthermore, this shared mechanism facilitates the construction of a unified feature representation, enabling the network to leverage critical information across tasks. Consequently, feature cross-modal matching can benefit from the restoration capabilities of $\bm{\hat{P}}$, enhancing the quality of the matched features.

\subsection{Omni Unified Feature Representation}
Due to the significant modality differences among various types of images, the features intended for alignment exhibit strong inconsistencies. This inconsistency directly impacts cross-modal feature matching. To address this, we leverage $\bm{h}_1$, $\bm{h}_2$, and $\bm{h}_3$, which integrate information from the source images. By incorporating them into the process of learning consistent features, we can compensate for the information loss in corresponding modality features. This not only reduces the differences among the features to be aligned but also enhances their representation power.
To this end, we design the OUFR module. In this module, to adapt to the feature input requirements of Spatial Mamba, we adopt the patch operation from \cite{32} to patchify ${\bm{\hat{F}'}_D}$ and ${\bm{F}_R}$, resulting in ${\hat{\bm{F}}'_D} = [\hat{\bm{f}}_D^{'1}, \hat{\bm{f}}_D^{'2}, \cdots, \hat{\bm{f}}_D^{'N}]$ and ${\hat{\bm{F}}_R} = [\hat{\bm{f}}_R^{1}, \hat{\bm{f}}_R^{2}, \cdots, \hat{\bm{f}}_R^{N}]$.

For computational convenience, we use  
$[\hat{\bm{f}}_{D}^{'1}, \hat{\bm{f}}_{D}^{'2}, \dots, \hat{\bm{f}}_{D}^{'N}, \hat{\bm{f}}_{R}^{1}, \hat{\bm{f}}_{R}^{2}, \dots, \hat{\bm{f}}_{R}^{N}]$ along with $\bm{h}_1$, $\bm{h}_2$, and $\bm{h}_3$ as inputs to Spatial Mamba. The output of the SSM in the $j$-th Spatial Mamba is given by:  
\begin{equation}\small  
	\hat{\bm{f}}_x^{ij} = {\rm{SSM}}(\hat{\bm{f}}_x^{(i,j-1)}, \bm{h}_{j}) = \bm{C}(\bar{\bm{A}} \bm{h}_j + \bar{\bm{B}} \hat{\bm{f}}_x^{(i,j-1)}),  
\end{equation}  
where $x=R,D$,  $j=1,2,3$, $\hat{\bm{f}}_x^{(1,0)}=\hat{\bm{f}}_x^{1}$, $\bar{\bm{A}}$ is the state selection matrix, and $\bar{\bm{B}}$ and $\bm{C}$ are modulation matrices obtained through linear transformations of $\hat{\bm{f}}_x^{(i,j-1)}$. Compared to standard Mamba, Spatial Mamba not only establishes relationships between different modal features through $\bm{h}_1$, $\bm{h}_2$, and $\bm{h}_3$ but also mitigates information loss in hidden features caused by forgetting in the SSM. We then separate the output of the third Spatial Mamba to obtain ${\bar{\bm{F}}'_D} \in \mathbb{R}^{N \times M}$ and ${\bar{\bm{F}}_R} \in \mathbb{R}^{N \times M}$.

To effectively eliminate the modality differences between ${\bar{\bm{F}}_D^{'}}$ and ${\bar{\bm{F}}_R}$, we introduce another image pair $\{\bm{I}_D, \bm{I}_R^{'}\}$. Here, $\bm{I}_D$ serves as the label of $\bm{I}_D^{'}$ aligned with $\bm{I}_R$, and $\bm{I}_R$ serves as the label of $\bm{I}_R^{'}$ aligned with $\bm{I}_D$, with neither $\bm{I}_D$ nor $\bm{I}_R^{'}$ exhibiting quality degradation. The pair $\{ \bm{I}_D, \bm{I}_R^{'} \}$ is processed through a shared encoder and Spatial Mamba to obtain ${\bar{\bm{F}}_D}$ and ${\bar{\bm{F}}_D^{'}}$. For ${\bar{\bm{F}}_D}$ and ${\bar{\bm{F}}_D^{'}}$, we useA ${\cal L}_{moda}$ to update the network parameters:
\begin{equation}\small 
	{\cal L}_{moda} = {\cal L}_{cont}({\bar{\bm{F}}_D},{\bar{\bm{F}}_R}) + {\cal L}_{cont}({\bar{\bm{F}}_R'},{\bar{\bm{F}}_D'}),
\end{equation}
where
\begin{equation}
	\resizebox{0.9\hsize}{!}{$
		\begin{aligned}
			&{\cal L}_{cont}(\bm{F}_1, \bm{F}_2) =\\
			& - \log \left( \frac{\exp\left({{\rm{sim}}(\bm{f}_i, \bm{f}_j^+)}/{\tau} \right)}
			{\exp\left( {{\rm{sim}}(\bm{f}_i, \bm{f}_j^+)}/{\tau} \right) + \sum_{j=1}^{K} \exp\left({{\rm{sim}}(\bm{f}_i, \bm{f}_j^-)}/{\tau} \right)} \right)
		\end{aligned}
		$}
\end{equation}
$\bm{f}_i$ and $\bm{f}_j^+$ are feature vectors sampled from the same position in $\bm{F}_1$ and $\bm{F}_2$, $\bm{f}_j^-$ represents one of $K$ feature vectors randomly sampled from other positions in $\bm{F}_2$, ${\rm{sim}}(\cdot,\cdot)$ denotes the cosine similarity, and $\tau = 0.1$ is the temperature coefficient.

\begin{figure}[t!]
	\centering
	\includegraphics[height=1.7in,width=2.9in]{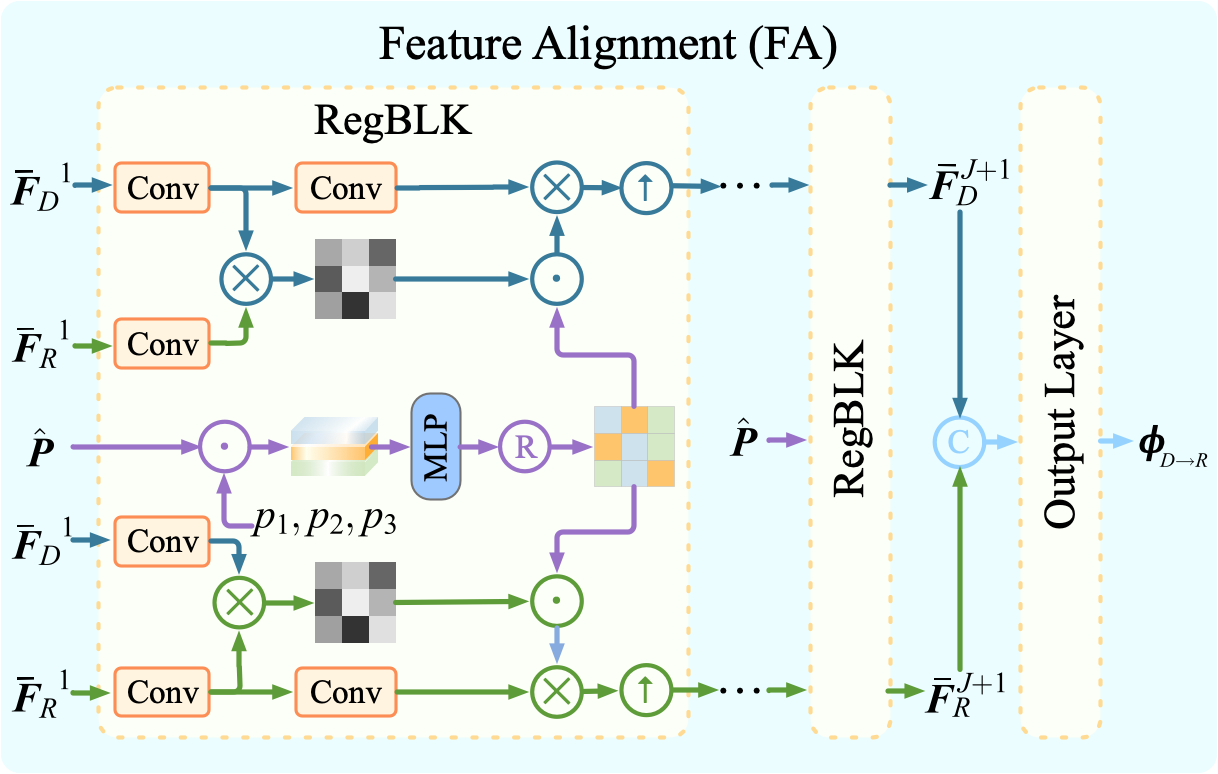}
	\caption{ Network architecture of the FA. }\vspace{-4.5mm}
	\label{fig3}
\end{figure}

\subsection{Feature Alignment}
After extracting the features $\bar{\bm{F}}_D^{'}$ and $\bar{\bm{F}}_R$, they are fed into the FA module to predict the deformation field $\bm{\phi}_{R \to D}$. The network architecture of the FA is illustrated in Figure \ref{fig3}. Specifically, $\bar{\bm{F}}_D^{'}$ and $\bar{\bm{F}}_R$ are spatially rearranged to obtain $\bar{\bm{F}}_D^1 \in \mathbb{R}^{M \times H/Q \times W/Q}$ and $\bar{\bm{F}}_R^1 \in \mathbb{R}^{M \times H/Q \times W/Q}$, where $Q$ represents the side length of the patch. Next, $\bar{\bm{F}}_D^1$ and $\bar{\bm{F}}_R^1$ are fed into a feature extraction block comprising $J$ RegBLKs to model spatial relationships between them. The $i$-th RegBLK takes $\bar{\bm{F}}_D^i$, $\bar{\bm{F}}_R^i$, and the degradation prompt $\hat{\bm{P}}$ as inputs, producing the outputs $\bar{\bm{F}}_D^{i+1}$ and $\bar{\bm{F}}_R^{i+1}$. In Figure \ref{fig3}, where $\uparrow$ denotes upsampling, MLP stands for multilayer perceptron, Conv represents a convolutional layer, and $\otimes$ indicates matrix multiplication. After passing through $J$ RegBLKs, $\bar{\bm{F}}_D^{J+1}$ and $\bar{\bm{F}}_R^{J+1}$ are concatenated and processed by an output layer consisting of a $3 \times 3 \times 3$ convolutional layer, a $1 \times 1 \times 1$ convolutional layer, and ReLU, to generate $\bm{\phi}_{D \to R}$. To enhance accuracy, we regularize the FA output using the deformation field label $\bm{\phi}_{gt}$:  
\begin{equation}
	\mathcal{L}_{reg} = \left\| \bm{\phi}_{gt} - \bm{\phi}_{D \to R} \right\|_F^2.
\end{equation}

\subsection{Universal Feature Restoration and Fusion}
%Reference image-guided restoration leverages a reference image to recover the quality of a degraded image. The reference image not only guides restoration but also injects useful information into the degraded image. This process shares similarities with image fusion, making their integration within a unified framework feasible. However, coordinating multiple degradation restoration tasks is challenging. To address this, we propose a multi-task joint feature representation method based on low-rank representation. Specifically, we design the Adaptive LoRA Synergistic Network (ALSN), which adaptively processes features from different degradation types without significantly increasing the parameter scale, enabling all-in-one restoration and fusion.

Reference image-guided restoration enhances degraded images by providing guidance and injecting useful information. This process aligns with image fusion, enabling their integration within a unified framework. However, coordinating multiple degradation restoration tasks is challenging. To address this, we propose a multi-task joint representation method based on low-rank representation. Specifically, we design the Adaptive LoRA Synergistic Network (ALSN), which adaptively processes features from different degradation types without significantly increasing the parameter scale, enabling all-in-one restoration and fusion.

As shown in Figure \ref{fig2}, we first obtain the input $\tilde{\bm{F}}$ for UFR\&F:
$\tilde{\bm{F}} = \text{Concat}(\mathcal{W}(\bm{F}_D^{'}, \bm{\phi}_{D \to R}), \bm{F}_R)$, 
where $\mathcal{W}$ denotes the warp operation. UFR\&F is a multi-scale U-shaped network comprising a contracting path on the left, an expanding path on the right, and skip connections in between. The ALSN is embedded in the skip connections and plays a key role in feature restoration and fusion. When $\tilde{\bm{F}}$ is fed into UFR\&F, the contracting path encodes it into multi-scale features. We denote the feature at the $i$-th scale as $\tilde{\bm{F}}_i$. Passing $\tilde{\bm{F}}_i$ through the corresponding ALSN module at that scale yields the restored and fused feature $\tilde{\bm{F}}_{i, rf}$. These features are then integrated and reconstructed into the final fused image $\bm{I}_{fuse}$ via the expanding path. Considering the differences between FA and feature restoration, we employ an MLP with parameters different from those used in FA to extract guidance information relevant to image feature restoration from $\hat{\bm{P}}$.

The computation of ALSN can be expressed as:
\begin{equation}\small
	\tilde{\bm{F}}_{i,rf} = {\cal B}_i( \tilde{\bm{F}}_i, \hat{\bm{P}}) + \sum_{j=1}^{\tilde{N}} p_j {\cal R}_{ij}( \tilde{\bm{F}}_i, \hat{\bm{P}}),
\end{equation}
where ${\cal B}_i$ represents the BaseNet at the $i$-th scale, and ${\cal R}_{ij}$ represents the $j$-th LoRANet at the $i$-th scale. This formulation arises because, in LoRA, the forward propagation of the network transforms from $\bm{y} = \bm{W}\bm{x}$ to  
$\bm{y} = \bm{W}'\bm{x} = \bm{W}\bm{x} + \bm{A}\bm{B}\bm{x}$. Extending this transformation to multiple branches, we obtain:
\begin{equation}\small
	\bm{y} = \bm{W}\bm{x} + \sum_{i=1}^{\tilde{M}} \lambda_i \bm{A}_i \bm{B}_i \bm{x}.
\end{equation}
By examining ${\cal B}_i$ in Eq.(7) and $\bm{W}$ in Eq. (8), we observe that both belong to the base network, allowing us to establish a correspondence between them. Similarly, ${\cal R}_{ij}$ in Eq. (7) and $\bm{A}_i \bm{B}_i$ in Eq. (8) belong to the LoRA branches, establishing a correspondence between ${\cal R}_{ij}$ and $\bm{A}_i \bm{B}_i$. Therefore, the computation of ALSN can be represented in the form of Eq. (8).

\begin{figure*}[ht!]
	\centering
	\includegraphics[height=2.5in,width=6.7in]{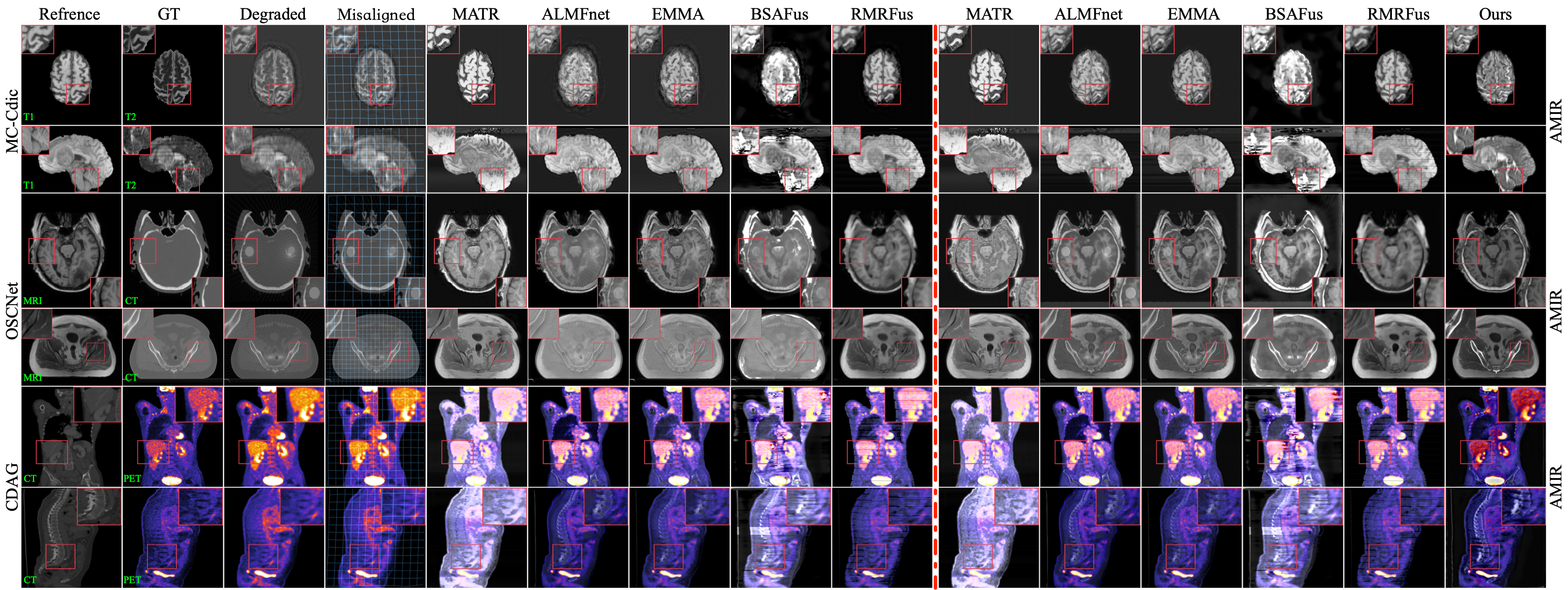}
	\caption{Visual Comparison of Fusion Results. The first column shows reference images without degradation, and the second column shows high-quality labels. The third and fourth columns present misaligned degraded images and their distortions. Columns 5 to 15 display results from different fusion methods. The image restoration methods to the right of the red line are shown on the right side of the figure, while those to the left are shown on the left side. }\vspace{-4.5mm}
	\label{fig4}
\end{figure*}
	
In the aforementioned method, relevant information beneficial to the current task can be selected from $\hat{\bm{P}}$ based on the image degradation category for different restoration tasks. Combined with the adaptive feature expression capabilities of each LoRA branch in ALSN, the network can select the most suitable parameters for the current task. Therefore, the proposed method effectively coordinates the different feature requirements posed by various tasks within a single framework. Meanwhile, the introduction of the LoRA network significantly reduces the number of model parameters. For ${{\cal B}_i}$, its inputs are ${\tilde{\bm{F}}_i}$ and $\hat{\bm{P}}$, and its output is $\tilde{\bm{F}}_i^B$. The input features first undergo channel attention:
\begin{equation}\small
	\bm{M}_i = {\rm{Resh}}( {\rm{MLP}}( \hat{\bm{P}} \odot \bm{p})) \odot {\rm{Conv}}( {\tilde{\bm{F}}_i}) \otimes {\rm{Conv}}( {\tilde{\bm{F}}_i}),
\end{equation}
where $\rm{Resh}$ represents the reshape operation. Then, forward propagation is performed on $\bm{V}_i^{'}=\bm{M}_i\otimes {\rm{Conv}}( {\tilde{\bm{F}}_i})$:
\begin{equation}\small
	\tilde{\bm{F}}_i^B = {\rm{Conv}}( {\bm{V}_i^{'}}) \odot {\rm{GELU}}( {\rm{Conv}}( {\bm{V}_i^{'}})).
\end{equation}
	
To maintain structural consistency between the fused image and the source images, we update the network parameters using a structural similarity loss:
\begin{equation}
	\resizebox{0.88\hsize}{!}{$
		{{\cal L}_{ssim}} = {\rm{SSIM}}( {\bm{I}_{fuse}, {\cal W}( {\bm{I}^{'}_{fine}, \bm{\phi}_{D \to R}})}) + {\rm{SSIM}}( {\bm{I}_{fuse}, \bm{I}_R}),
		$}
\end{equation}
where SSIM denotes the structural similarity, and ${\bm{I}^{'}_{fine}}$ is the high-quality image label corresponding to the degraded image ${\bm{I}^{'}_{D}}$. To enhance the contrast and preserve sharp edge details in the fused image, we introduce pixel intensity loss and gradient loss:
\begin{equation}
	\resizebox{0.88\hsize}{!}{$
		\begin{aligned}
			{{\cal L}_{inten}} &= \left\| \bm{I}_{fuse} - 
			\max \Big( {\cal W} ( \bm{I}^{'}_{fine}, \bm{\phi}_{D \to R} ), \bm{I}_R \Big) \right\|_1, \\
			{{\cal L}_{grad}} &= \left\| \nabla \bm{I}_{fuse} - 
			\max \Big( \nabla {\cal W} ( \bm{I}^{'}_{fine}, \bm{\phi}_{D \to R} ), \nabla \bm{I}_R \Big) \right\|_1,
		\end{aligned}
		$}
\end{equation}
where $\nabla$ represents the gradient operator. 

To ensure accurate color reconstruction and preservation when processing color images, in addition to the fusion loss for the Y channel, we supervise the CrCb channels using a color loss:
\begin{equation}\small
	{{\cal L}_{crcb}} = {\left\| {\bm{C}_{fuse} - \bm{C}_{fine}} \right\|_1} + {\rm{SSIM}}\left( {\bm{C}_{fuse}, \bm{C}_{fine}} \right),
\end{equation}
where ${\bm{C}_x} (x = fuse, fine)$ represents the CrCb channels of the corresponding images. % Note that ${{\cal L}_{crcb}} = 0$ when the input image is single-channel grayscale. 
The overall loss is defined as:
\begin{equation}\small
	{{\cal L}_{rf}} = {{\cal L}_{ssim}} + {{\cal L}_{inten}} + {{\cal L}_{grad}} + {{\cal L}_{crcb}}.
\end{equation}
Thus, the total loss function in this study is:
\begin{equation}\small
	{\cal L} = {{\cal L}_{ce}} + {{\cal L}_{moda}} + {{\cal L}_{reg}} + {{\cal L}_{rf}}.
\end{equation}	
 \vspace{-4.5mm}
\section{Experiments}
\subsection{Construction of Training and Testing Sets}
In this paper, we evaluate our model on three medical image fusion tasks involving degraded images: (1) fusion of high-quality MRI-T1 with MRI-T2 affected by motion artifacts, (2) fusion of high-quality MRI with CT degraded by metal artifacts and noise, and (3) fusion of high-quality CT with low-dose PET affected by noise. For the first task, we use the BraTs2020 \cite{33} dataset, which includes 494 pairs of strictly aligned high-quality MRI-T1 and MRI-T2 images, split into 369 pairs for training and 125 for testing. We synthesize MRI-T2 images with motion artifacts using the method described in \cite{34}. For the second task, we use the SynthRAD2023 \cite{35} dataset, containing 360 pairs of aligned high-quality MRI and CT images. We randomly select 298 pairs for training and 62 for testing, and generate degraded CT images using the method in  \cite{36}. For the third task, we use the FDG-PET/CT \cite{37} dataset, which includes 2939 pairs of aligned high-quality CT and PET images. We select 500 pairs for testing and use the rest for training, synthesizing degraded PET images following  \cite{38}. To simulate misalignment, we apply rigid and non-rigid transformations to the degraded images.

\vspace{-1.0mm}
\subsection{Implementation Details}\vspace{-1.0mm}
We train the model end-to-end with a batch size of 3 for 500 epochs. Each image is uniformly resized to $160 \times 160 \times 160$, with random cropping and scaling for data augmentation. Model parameters are optimized using AdamW with an initial learning rate of $1 \times 10^{-4}$. A cosine annealing schedule gradually reduces the learning rate to $1 \times 10^{-6}$ by the final epoch for stable training. Gradient clipping is applied to keep the $l_2$-norm of gradients below 1. The method is implemented in PyTorch and trained on a single NVIDIA GeForce RTX 4090 GPU.\vspace{-1.5mm}
\subsection{Comparison Methods and Evaluation Metrics}\vspace{-1.0mm}
Since the method proposed is the first designed for the fusion of degraded and misaligned medical images, there are no directly comparable methods. To validate its effectiveness, we adopt a multi-stage processing paradigm: Restoration, Registration, and Fusion, as a baseline for comparison. In the Restoration stage, we select the MRI artifact removal method MC-CDic~\cite{5}, the CT artifact and noise reduction method OSCNet~\cite{23}, the PET denoising method CDAG~\cite{12}, and the All-in-One medical image restoration method AMIR~\cite{13} to restore the degraded images before fusion. In the Registration stage, we use the high-performance 3D medical image registration method CorrMLP~\cite{39}. Finally, in the Fusion stage, we apply state-of-the-art multimodal medical image fusion techniques, including MATR~\cite{3}, ALMFnet~\cite{40}, BSAFus~\cite{27}, EMMA~\cite{21}, and RMRFus~\cite{22}, to generate the final fused results. To objectively evaluate the performance of the algorithm, we adopt six commonly used evaluation metrics: Mean Squared Error ($Q_{mse}$)~\cite{41}, Correlation Coefficient ($Q_{cc}$)~\cite{42}, Peak Signal-to-Noise Ratio ($Q_{psnr}$)~\cite{43}, Structural Content Difference ($Q_{scd}$)~\cite{41}, Visual Information Fidelity ($Q_{vif}$)~\cite{45}, and Structural Similarity Index ($Q_{ssim}$)~\cite{46}. Additionally, we compare the computational complexity of the models using FLOPs ($Q_{f}$) metric, where lower values of $Q_{mse}$ and $Q_{f}$ indicate better performance, while higher values for the other metrics are preferable.

\begin{table}[t!]
	\centering
\caption{Performance Comparison on BraTS2020. ``M'' denotes MC-Cbic, ``A'' denotes AMIR.%, and bold indicates the best performance.
} \vspace{-2.5mm}
	\resizebox{\columnwidth}{!}{%
		\begin{tabular}{l|cccccc}
			\toprule
			Methods & {$Q_{mse} \downarrow$} & {$Q_{cc} \uparrow$} & {$Q_{psnr} \uparrow$} & {$Q_{vif} \uparrow$} & {$Q_{ssim} \uparrow$} & {$Q_{f}(G) \downarrow$} \\
			\midrule
			M+MATR & 0.0458 & 0.8139 & 13.4179 & 1.2805 & 0.3831 & 184,069 \\
			M+ALMFnet & 0.0222 & 0.8970 & 16.7942 & 1.5211 & 0.2918 & 186,567 \\
			M+EMMA & 0.0243 & 0.8812 & 15.9361 & 1.1290 & 0.2710 & 185,838 \\
			M+BSAFus & 0.0397 & 0.8008 & 14.0658 & 1.7141 & 0.5484 & 186,997 \\
			M+RMRFus & 0.0215 & 0.8843 & 17.4663 & 1.0162 & 0.6121 & 190,071 \\
			\midrule
			A+MATR & 0.0464 & 0.8124 & 13.3615 & 1.2844 & 0.3836 & 31,920 \\
			A+ALMFnet & 0.0252 & 0.8940 & 16.1867 & 1.2850 & 0.2750 & 34,418 \\
			A+EMMA & 0.0234 & 0.8962 & 16.0901 & 0.9927 & 0.2629 & 33,689 \\
			A+BSAFus & 0.0414 & 0.7910 & 13.8845 & 1.4192 & 0.5386 & 34,848 \\
			A+RMRFus & 0.0188 & 0.8802 & 17.2326 & 0.9986 & 0.5686 & 37,922 \\
			\midrule
			\textbf{Ours} & \textbf{0.0125} & \textbf{0.9082} & \textbf{23.0727} & \textbf{1.9157} & \textbf{1.5288} & \textbf{395} \\
			\bottomrule
	\end{tabular}}\vspace{-4.5mm}
	\label{table1}
\end{table}
\vspace{-1.5mm}
\subsection{Experimental Result Analysis}\vspace{-1.5mm}
\textbf{Comparison of Results on BraTS2020}. 
The MRI-T2 imaging process is often time-consuming, during which slight patient movement may introduce inevitable motion artifacts. To assess our method's performance in image fusion when source images contain motion artifacts, we compare it against a multi-stage combination strategy on the BraTs2020 dataset. For the comparison method, we first restore the degraded MRI-T2 images using MC-CDic and AMIR, respectively. Next, we align the restored images with MRI-T1 images using CorrMLP. Finally, we apply five image fusion techniques to generate the final fused images. 
The top two rows of Figure~\ref{fig4} present the fusion results of different methods. It can be observed that our method demonstrates significant advantages in spatial alignment, motion artifact elimination, contrast enhancement, and detail preservation. Table~\ref{table1} presents a quantitative comparison of each method's performance on the dataset. The results indicate that our method achieves the highest average score on all five fusion metrics while significantly reducing computational cost compared to other methods. 
%Notably, our method performs best in terms of SSIM, primarily benefiting from its integrated approach to feature alignment, restoration, and fusion, which preserves the overall structural consistency of the data. In contrast, the multi-stage combination method processes each stage independently, preventing mutual reinforcement between tasks and making it difficult to achieve optimal performance. Additionally, this method handles 3D images by slicing them into 2D images, disrupting continuity between slices and leading to a decline in performance.
\begin{table}[t!]
	\centering
		\caption{Performance Comparison on SynthRAD2023 Dataset. ``O'' denotes OSCNet, ``A'' denotes AMIR. %, and bold indicates the best performance.
		}\vspace{-2.5mm}
	\resizebox{\columnwidth}{!}{%
		\begin{tabular}{l|cccccc}
			\toprule
			Methods & {$Q_{mse} \downarrow$} & {$Q_{cc} \uparrow$} & {$Q_{psnr} \uparrow$} & {$Q_{vif} \uparrow$} & {$Q_{ssim} \uparrow$} & {$Q_{f}(G) \downarrow$} \\
			\midrule
			O+MATR & 0.0496 & 0.7840 & 13.1633 & 1.4037 & 0.3112 & 20,891 \\
			O+ALMFnet & 0.0121 & 0.7852 & 17.6213 & 1.0720 & 0.5476 & 23,389 \\
			O+EMMA & 0.0215 & 0.7952 & 18.5446 & 1.0200 & 1.2168 & 23,819 \\
			O+BSAFus & 0.0412 & 0.7673 & 13.9823 & 1.2716 & 0.6152 & 22,660 \\
			O+RMRFus & 0.0123 & 0.7892 & 18.7646 & 0.9880 & 1.0738 & 26,893 \\
			\midrule
			A+MATR & 0.0540 & 0.7623 & 12.8122 & 1.3121 & 0.2816 & 31,920 \\
			A+ALMFnet & 0.0125 & 0.8004 & 18.9958 & 1.1466 & 0.7971 & 34,418 \\
			A+EMMA & 0.0138 & 0.8021 & 18.3619 & 1.0529 & 0.6544 & 34,848 \\
			A+BSAFus & 0.0618 & 0.6934 & 12.2731 & 1.3970 & 0.3154 & 33,689 \\
			A+RMRFus & 0.0125 & 0.7906 & 19.0780 & 0.9857 & 1.0020 & 37,922 \\
			\midrule
			\textbf{Ours}& \textbf{0.0118} & \textbf{0.8118} & \textbf{20.3548} & \textbf{1.4164} & \textbf{1.2414} & \textbf{395} \\
			\bottomrule
	\end{tabular}}\vspace{-4.5mm}
	\label{table2}
\end{table}
\begin{table}[ht!]
	\centering
	\caption{Performance Comparison on  FDG-PET/CT Dataset.``C'' denotes CDAG, ``A'' denotes AMIR.%, and bold indicates the best performance.
	}\vspace{-2.5mm}
	\resizebox{\columnwidth}{!}{%
		\begin{tabular}{l|cccccc}
			\toprule
			Methods & {$Q_{mse} \downarrow$} & {$Q_{cc} \uparrow$} & {$Q_{psnr} \uparrow$} & {$Q_{vif} \uparrow$} & {$Q_{ssim} \uparrow$} & {$Q_{f}(G) \downarrow$} \\
			\midrule
			C+MATR & 0.0349 & 0.8561 & 14.6079 & 0.8958 & 0.9554 & 13,061 \\
			C+ALMFnet & 0.0160 & 0.8587 & 18.1769 & 0.6607 & 0.9420 & 15,559 \\
			C+EMMA & 0.0123 & 0.8668 & 19.2313 & 0.8402 & 0.9585 & 15,989 \\
			C+BSAFus & 0.0290 & 0.7922 & 15.5364 & 0.9873 & 0.5666 & 14,830 \\
			C+RMRFus & 0.0171 & 0.8445 & 20.5325 & 0.9900 & 0.6099 & 19,063 \\
			\midrule
			A+MATR & 0.0341 & 0.8427 & 14.7088 & 0.9230 & 0.9535 & 31,920 \\
			A+ALMFnet & 0.0141 & 0.8576 & 18.8114 & 0.8279 & 0.8180 & 34,418 \\
			A+EMMA & 0.0102 & 0.8505 & 20.1927 & 0.9020 & 0.9596 & 34,848 \\
			A+BSAFus & 0.0279 & 0.7809 & 15.7322 & 0.8865 & 0.5728 & 33,689 \\
			A+RMRFus & 0.0150 & 0.8612 & 20.0572 & 1.0337 & 0.7500 & 37,922 \\
			\midrule
			\textbf{Ours} & \textbf{0.0057} & \textbf{0.8710} & \textbf{23.0076} & \textbf{1.1054} & \textbf{0.9716} & \textbf{395} \\
			\bottomrule
	\end{tabular}}\vspace{-3.5mm}
	\label{table3}
\end{table}

\textbf{Comparison of Results on SynthRAD2023}. 
Metal implants or foreign objects are commonly found in patients, which inevitably introduce metal artifacts in CT images, degrading fusion quality. To assess our method’s fusion performance on images degraded by metal artifacts, we performed comparisons on the SynthRAD2023 dataset. Specifically, we first restored the CT images affected by metal artifacts using OSCNet and AMIR. Then, following the same processing pipeline as for the BraTs2020 dataset, we employed CorrMLP for image registration. Finally, we applied five fusion methods to generate the final fused images.
The third and fourth rows of Figure~\ref{fig4} illustrate the visual fusion results of each method on this dataset. As observed, our method effectively preserves structural integrity and contrast while mitigating feature mismatches and metal artifacts. Table~\ref{table2} presents a quantitative comparison, showing that our method outperforms all others across all metrics while maintaining significantly lower computational cost.
\begin{table}[t!]
	\centering
	\caption{Ablation study on BraTs2020 Dataset.}\vspace{-2.5mm}
	\renewcommand{\arraystretch}{1.1}%
	\resizebox{0.425\textwidth}{!}{%
		\begin{tabular}{c|ccccc}
			\hline
			\multirow{2}{*}{Settings} & \multicolumn{5}{c}{Metrics} \\ 
			\cline{2-6}
			& $Q_{mse} \downarrow$ & $Q_{cc} \uparrow$ & $Q_{psnr} \uparrow$ & $Q_{vif} \uparrow$ & $Q_{ssim} \uparrow$ \\
			\hline
			Setting A & 0.0276 & 0.9021 & 21.5925 & 1.4905 & 1.0828  \\ 
			Setting B & 0.0187 & 0.8906 & 19.2927 & 1.7242 & 0.9577  \\ 
			Setting C & 0.0284 & 0.7360 & 17.0832 & 1.667 & 0.9146  \\ 
			\hline
			Transformer & 0.0229 & 0.8831 & 20.4635 & 1.5788 & 1.3128   \\ 
			Mamba & 0.0262 & 0.8736 & 21.1814 & 1.7048 & 1.2021   \\ 
			w/o OUFR & 0.0314 & 0.8633 & 20.0871 & 1.6842 & 1.1594   \\ 
			
			\hline
			w/o LoRA       & 0.0485          & 0.8937          & 21.1625         & 1.8053          & 1.4548          \\
			Multi-Experts  & 0.0605          & 0.8993          & 22.3880         & 1.7025          & 1.4400          \\
			
			\hline
			\textbf{Ours} & \textbf{0.0125} & \textbf{0.9082} & \textbf{23.0727} & \textbf{1.9157} & \textbf{1.5288} \\ 
			\hline
		\end{tabular}%
	}\vspace{-4.5mm}
	\label{table4}
\end{table}

\textbf{Comparison of Results on FDG-PET/CT}. 
To reduce radiation exposure in PET imaging, low-dose PET images often suffer from noise. To assess our method’s fusion performance on noisy PET images, we compared it with multi-stage approaches. Specifically, we first restored the degraded PET images using CDAG and AMIR, then performed registration and fusion following the same procedure as for the previous datasets. The fifth and sixth rows of Figure~\ref{fig4} present a visual comparison of different methods on this dataset, showing that our method significantly outperforms others in color restoration, noise reduction, feature alignment, and contrast enhancement. Table~\ref{table3} provides a quantitative evaluation on the FDG-PET/CT dataset, showing that our method achieves the best performance across all metrics. \textbf{More experimental results are provided in the supplementary material.}

\vspace{-1.5mm}
\subsection{Ablation Study}\vspace{-1.5mm}
In this section, we evaluate the effectiveness of the four core components in our model: DAPL, OUFR, FA, and UFR\&F.

\textbf{Effectiveness of DAPL}. To evaluate its effectiveness, we perform three comparative experiments under the following scenarios: (A) DAPL lacks comprehensive feature support in mitigating modality differences; (B) it does not supply degradation-related prompts to FA or, if applicable, the fusion-repair network; (C) DAPL is completely removed. As shown in Table~\ref{table4}, the absence of comprehensive feature support leads to a significant decline in fusion performance. Similarly, the absence of degradation-related prompts negatively impacts multi-task processing. The model reaches optimal performance only when DAPL is fully integrated, highlighting its effectiveness.

%\textbf{Effectiveness of DAPL}. The DAPL module is essential in two ways: (1) it helps eliminate modal differences, aiding feature alignment (FA) for feature merging and reducing the impact of image misalignment on fusion; (2) it facilitates task interactions through unified, learnable prompts, enhancing multi-task collaboration and influencing the final fusion outcome. To validate its effectiveness, we conducted three comparison experiments under three settings: (A) DAPL does not provide comprehensive feature support for eliminating modal differences; (B) it does not provide degradation-related prompts to the FA and, if applicable, the fusion-repair network; (C) it is entirely removed. Table 4 shows that without comprehensive feature assistance, fusion performance drops significantly. Similarly, without degradation-related prompts, multi-task processing suffers. Only when DAPL is fully utilized does the model achieve optimal performance, demonstrating its effectiveness.

\textbf{Effectiveness of OUFR}.
To validate the effectiveness of OUFR, we replaced the proposed Spatial Mamba with either a Transformer or the standard Mamba for comparison. Table \ref{table4} presents the ablation study results for OUFR, showing that Spatial Mamba outperforms both the Transformer and the standard Mamba. Furthermore, the best fusion performance is achieved only when the full OUFR module is integrated into the network. This fully confirms the effectiveness of OUFR.

\begin{table}[h]
	\centering
	\caption{Ablation study on FA: Performance comparison under different settings. ``N-J" indicates the number of RegBLKs.}\vspace{-2.5mm}
	\label{table5} % 确保 label 在 caption 之后
	\resizebox{0.38\textwidth}{!}{%
		\begin{tabular}{c|ccccc}
			\hline
			\multirow{2}{*}{N-J} & \multicolumn{5}{c}{Metrics} \\ 
			\cline{2-6}
			& {$Q_{mse} \downarrow$} & {$Q_{cc} \uparrow$} & {$Q_{psnr} \uparrow$} & {$Q_{vif} \uparrow$} & {$Q_{ssim} \uparrow$} \\
			\hline
			0 & 0.0489 & 0.7050 & 17.0852 & 0.9795 & 0.8326 \\ 
			2 & 0.0214 & 0.8369 & 20.1934 & 1.2185 & 0.9015 \\ 
			3 & 0.0178 & 0.8529 & 22.3577 & 1.3479 & 1.1375 \\ 
			4 & 0.0100 & \textbf{0.8637} & \textbf{23.0727} & \textbf{1.4792} & \textbf{1.2473}  \\ 
			5 & \textbf{0.0098} & 0.8611 & 22.8901 & 1.3820 & 1.2185 \\ 
			\hline
		\end{tabular}%
	}\vspace{-3.5mm}
\end{table}

\textbf{Effectiveness of FA}.
The proposed FA module aligns the fused features using the predicted deformation field, thereby reducing the adverse effects of misaligned source images on fusion quality. To evaluate the effectiveness of incorporating multiple RegBLK modules in the FA module, we conduct an ablation study by adjusting the number of RegBLKs. As shown in Table \ref{table5}, the network achieves optimal performance when the number of RegBLKs ($J$) is set to 4.
\begin{figure}[t!]
	\centering
	\includegraphics[height=1.9in,width=2.5in]{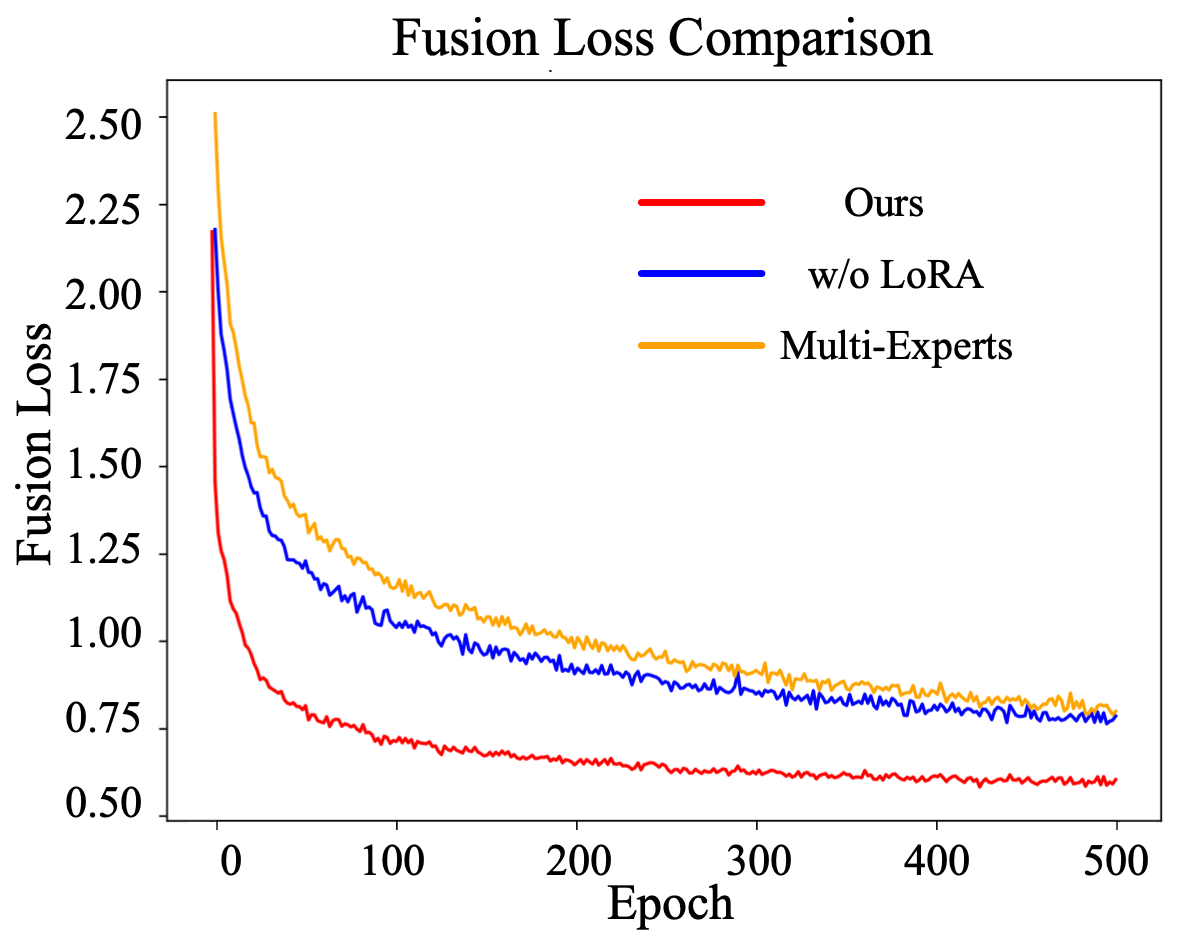}
	\caption{ Ablation study on UFR\&F: Fusion loss reduction under different settings.
		“w/o LoRA” removes all LoRA branches, while “Multi-Expert” replaces ALSN with a traditional multi-expert architecture. }\vspace{-4.5mm}
	\label{fig5}
\end{figure}

\textbf{Effectiveness of UFR\&F}.
To validate the effectiveness of UFR\&F, we conduct two controlled experiments. In the first, we remove the LoRA branch from ALSN, retaining only BaseNet. In the second, we replace the LoRA branch with a multi-expert structure consisting of multiple BaseNet instances. As shown in Table~\ref{table4}, removing LoRA significantly reduces performance due to the loss of task-specific adaptability. Replacing it with a multi-expert structure also degrades performance as shown in Figure \ref{fig5}, as the increased parameter scale hinders convergence. Statistical analysis indicates that ALSN reduces the parameter scale of UFR\&F by 68.22\%, from 2.986M to 0.948M. These results validate the effectiveness of UFR\&F.\textbf{ More experimental results are provided in the supplementary material}.\vspace{-1.5mm}

%To validate the effectiveness of UFR\&F, we designed two controlled experiments. In the first, we removed the LoRA branch from ALSN, keeping only the BaseNet component. In the second, we replaced the LoRA branch with a traditional multi-expert structure composed of BaseNet. The results show that when UFR\&F retains only BaseNet, its multi-task performance significantly declines due to the loss of the network’s task-specific adaptability. Similarly, in the second experiment, performance also dropped, as the multi-expert architecture increased the parameter scale, hindering model convergence and preventing it from reaching the expected performance within the same number of training epochs (see Figure.\ref{fig5}). Furthermore, statistics show that introducing ALSN reduces the parameter scale of UFR\&F by 68.22\% compared to the traditional multi-expert architecture, from 2.986M to 0.948M. These experiments fully demonstrate the effectiveness of UFR\&F.

\section{Conclusion}
We propose UniFuse, a general multimodal medical image fusion framework that addresses misalignment and degradation in source images. Unlike conventional methods, it integrates a degradation-aware prompt learning module to enhance cross-modal alignment and restoration. Our OUFR, powered by Spatial Mamba, mitigates modality differences and improves feature alignment. Meanwhile, the UFR\&F, incorporating ALSN, enables efficient joint restoration and fusion within a single-stage framework. Experiments across multiple datasets demonstrate the superiority of UniFuse over existing methods in handling degraded and misaligned medical images. Future work will focus on enhancing model adaptability and efficiency for broader clinical applications.

{
    \small
    \bibliographystyle{ieeenat_fullname}
    \bibliography{main}
}
\clearpage
\setcounter{page}{1}
\maketitlesupplementary
\renewcommand{\thesubsection}{\Alph{subsection}} 
\section{Further Experimental Comparisons}
\begin{figure}[ht!]
	\centering
	\includegraphics[height=1.80in,width=3.0in]{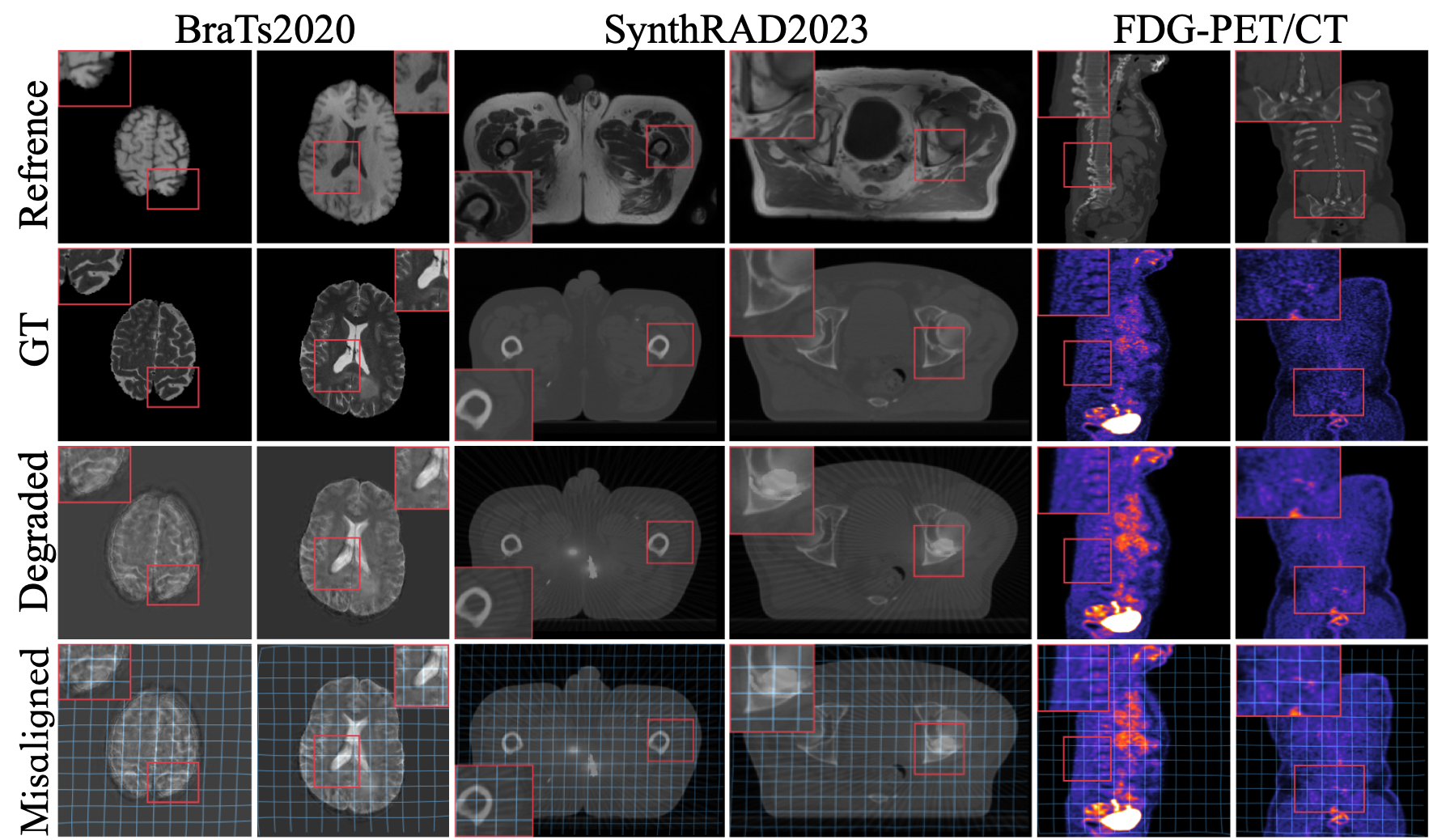}
	\caption{Source images to be fused and corresponding GTs of degraded images. The first row shows the reference images, while the second row presents the GTs of degraded images. The third and fourth rows display misaligned degraded images and their distortions.}\label{c7}
\end{figure}

\begin{figure}[ht!]
	\centering
	\includegraphics[height=4.80in,width=3.0in]{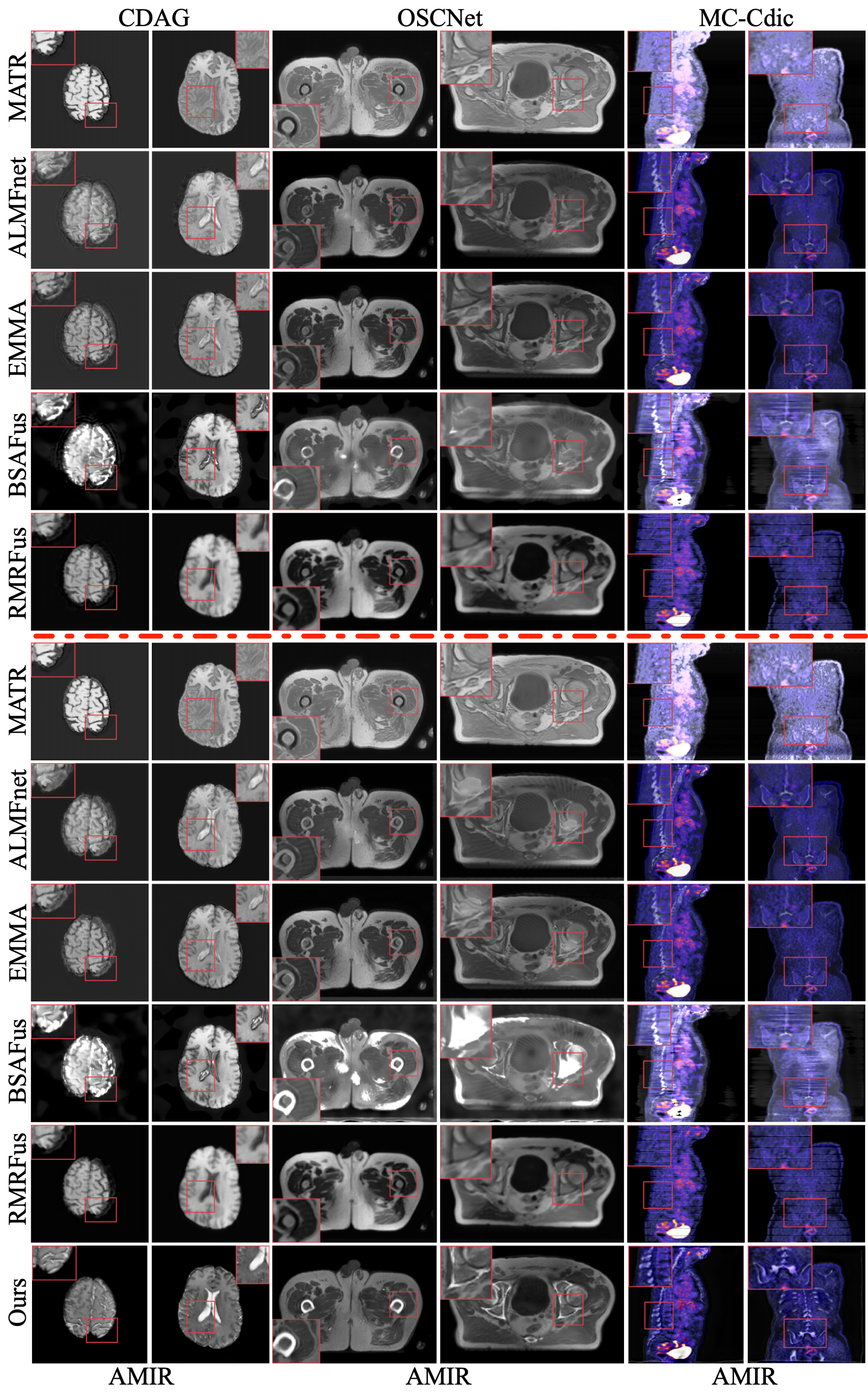}
	\caption{Comparison of visual effects of fusion results using different methods. Rows 1 to 11 display the results of various fusion methods. The text at the top of the image indicates the image restoration method used for the fusion results above the red line, while the text at the bottom indicates the image restoration method used for the fusion results below. The text on the left specifies the fusion method used for each row.}\label{c8}
\end{figure}

To comprehensively evaluate the performance of our proposed method, we provide additional visual comparisons of fusion results. Figure \ref{c7} presents source images from different datasets with various types of degradation. After processing these images using different fusion methods, the corresponding fusion results are shown in Figure \ref{c8}. As observed in Figure \ref{c8}, our method exhibits significant advantages in artifact suppression,  noise elimination, feature alignment, and the restoration of contrast and detail. These results further demonstrate the superiority of our method in terms of visual quality.

\section{Evaluation of Alignment Performance}
\label{sec:rationale} 
\begin{figure}[ht!]
	\centering
	\includegraphics[height=4.90in,width=3.0in]{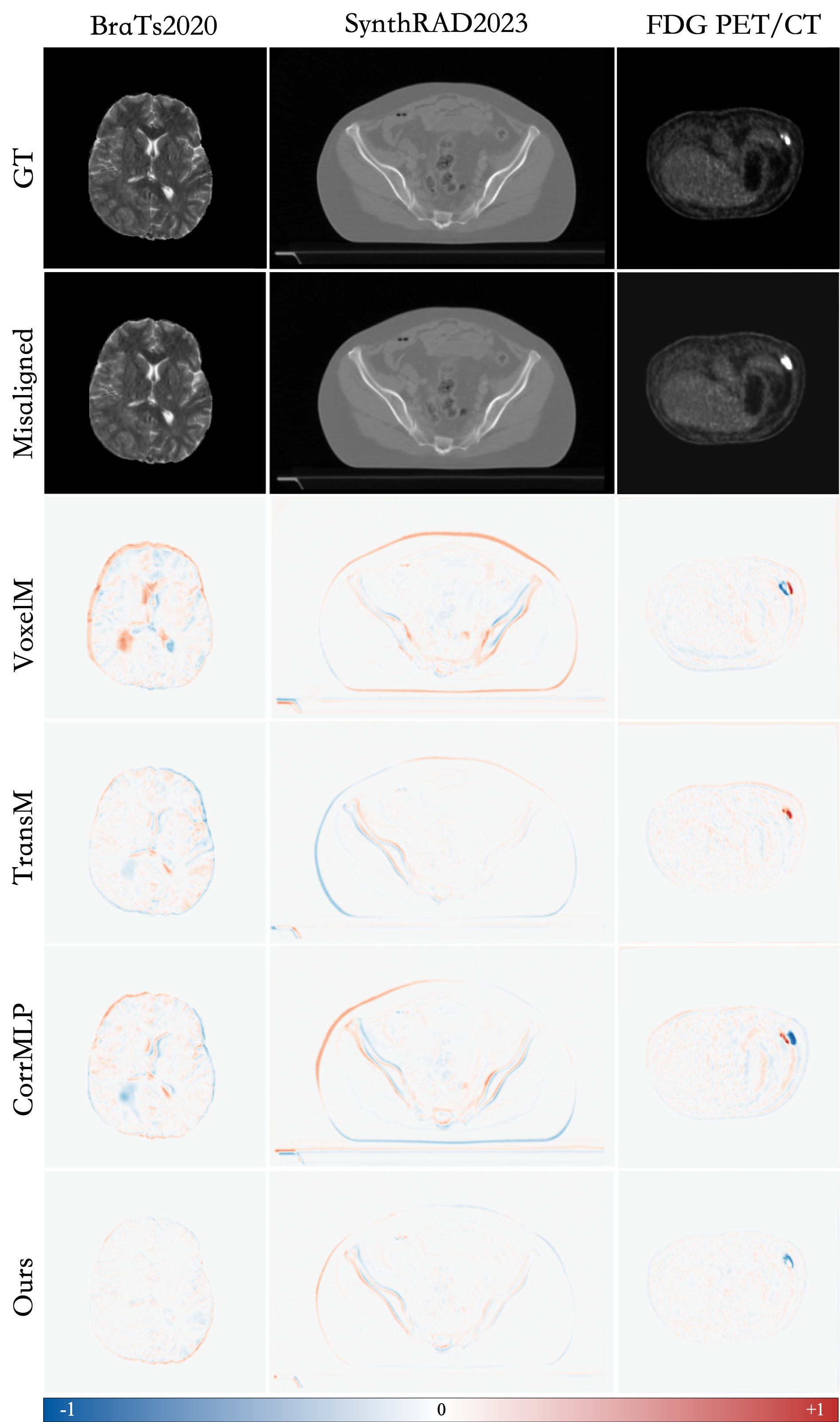}
	\caption{Visual comparison of alignment performance. The first row shows the aligned image labels, the second row presents the misaligned images, and the third to sixth rows display the alignment error maps for different methods. The closer the image information is to white, the better the alignment performance.}\label{c1}
\end{figure}

To evaluate the alignment performance of our method, we use the deformation field generated by the UniFuse network for image alignment and compare it with other image alignment methods, including VoxelM \cite{c1}, TransM \cite{c2}, and CorrMLP \cite{c3}. To quantify the alignment results, we adopt commonly used alignment metrics: mean squared error ($Q_{mse}$), correlation coefficient ($Q_{cc}$), and structural similarity ($Q_{ssim}$). To ensure a fair comparison, all unaligned degraded images are first processed by AMIR for image quality restoration before feeding them into the comparison methods. As evident from Table \ref{tabc1}, our method demonstrates significant overall advantages across different datasets. This is primarily attributed to the specially designed degradation-aware prompt module, which ensures robust feature alignment performance when handling various types of data.
To visually assess the quality of alignment, we generate alignment error maps by subtracting the aligned images from their corresponding label images. For better visualization, we render the error maps using the color spectrum shown below the figure, where perfectly aligned regions appear white. As clearly shown in the alignment error maps in Figure \ref{c1}, our method exhibits a significant advantage in feature alignment compared to other methods.

\begin{table*}[t]
	\centering
	\scriptsize 
	\renewcommand{\arraystretch}{1.1} 
	\setlength{\tabcolsep}{3pt}      
	\caption{Comparison of alignment performance across different methods and datasets.}
	\resizebox{\textwidth}{!}{
		\begin{tabular}{lccc|lccc|lccc}
			\toprule
			\multicolumn{4}{c|}{\textbf{BraTs2020 Dataset}} & \multicolumn{4}{c|}{\textbf{SynthRAD2023 Dataset}} & \multicolumn{4}{c}{\textbf{FDG PET/CT Dataset}} \\
			\cmidrule(lr){1-4} \cmidrule(lr){5-8} \cmidrule(lr){9-12}
			Methods & {$Q_{mse} \downarrow$} & {$Q_{cc} \uparrow$} & {$Q_{ssim} \uparrow$}  & Methods & {$Q_{mse} \downarrow$} & {$Q_{cc} \uparrow$} & {$Q_{ssim} \uparrow$}  & Methods & {$Q_{mse} \downarrow$} & {$Q_{cc} \uparrow$} & {$Q_{ssim} \uparrow$}  \\
			\midrule
			VoxelM & 1.49E-3 & 0.456 & 0.885 & VoxelM & 7.02E-4 & 0.485 & 0.902 & VoxelM & 3.01E-4 & 0.477 & 0.927 \\
			TransM & 5.07E-4 & 0.485 & 0.961 & TransM & 7.24E-4 & 0.484 & 0.906 & TransM & 2.92E-4 & 0.478 & 0.919 \\
			CorrMLP & 4.89E-4 & 0.485 & 0.962 & CorrMLP & 5.12E-4 & 0.490 & 0.931 & CorrMLP & 2.42E-4 & 0.482 & \textbf{0.968} \\
			Ours & \textbf{4.21E-4} & \textbf{0.488} & \textbf{0.966} & Ours & \textbf{3.48E-4} & \textbf{0.492} & \textbf{0.951} & Ours & \textbf{9.52E-5} & \textbf{0.493} & 0.935 \\
			\bottomrule
		\end{tabular}\label{tabc1}%
	}
\end{table*}

\section{Further Ablation Study}
\begin{figure*}[t!]
	\centering
	\includegraphics[height=2.50in,width=6.5in]{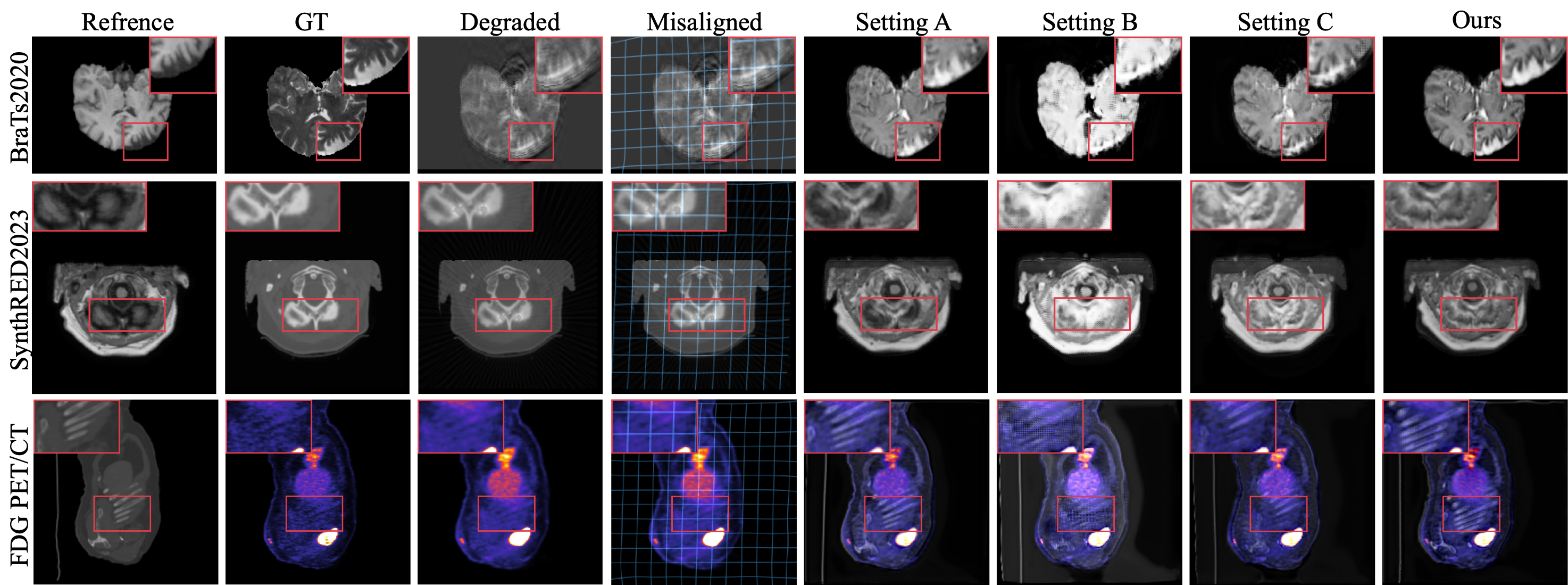}
	\caption{Visual comparison of DAPL's effectiveness.	The first column shows the reference image without degradation, the second column presents the ground truth (GT) of the degraded image, the third column displays the misaligned degraded image, the fourth column highlights the distortions in the misaligned degraded image, and the fifth to eighth columns show the fusion results under different experimental settings.}\label{c2}
\end{figure*}

\textbf{Effectiveness of DAPL}. To visually analyze the effectiveness of DAPL, we visualize the results under the three settings in the DAPL ablation study, as shown in Figure \ref{c2}. In Setting A, the fusion results of the network exhibit feature misalignment due to the lack of assistance from DAPL, which reduces the effectiveness of OUFR and subsequently affects the feature alignment performance of FA. In Setting B, the fusion results show significant contrast deviations and a loss of source image details due to the absence of degradation-aware prompts in the fusion and restoration network, making it difficult for the network to maintain consistent fusion performance across different degradation task scenarios. In Setting C, the fusion results display a decline in both feature alignment and detail restoration performance. Only when DAPL is fully retained do the fusion results achieve optimal visual quality.

\begin{figure*}[ht!]
	\centering
	\includegraphics[height=2.50in,width=6.5in]{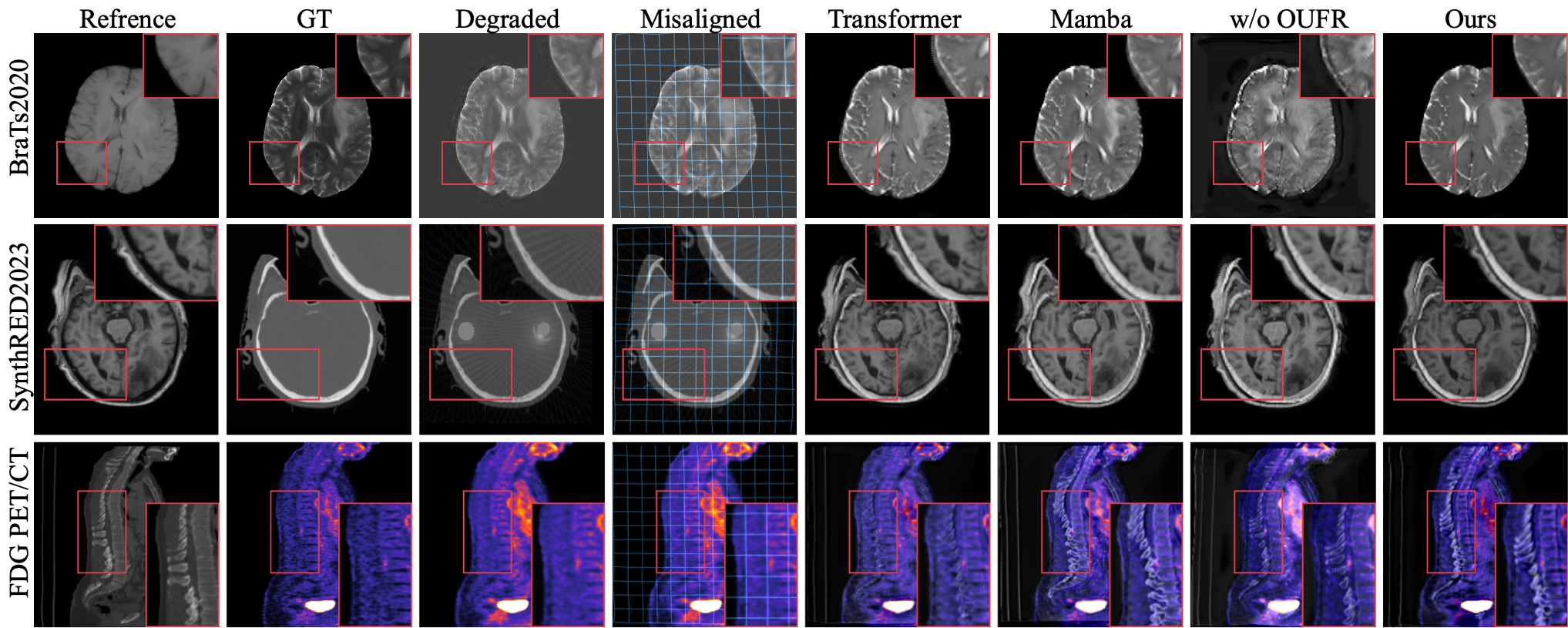}
	\caption{Visual comparison of OUFR's effectiveness.	The first column shows the reference image without degradation, the second column presents the ground truth (GT) of the degraded image, the third column displays the misaligned degraded image, the fourth column highlights the distortions in the misaligned degraded image, and the fifth to eighth columns show the fusion results under different experimental settings.}\label{c3}
\end{figure*}
\begin{figure*}[ht!]
	\centering
	\includegraphics[height=2.60in,width=6.6in]{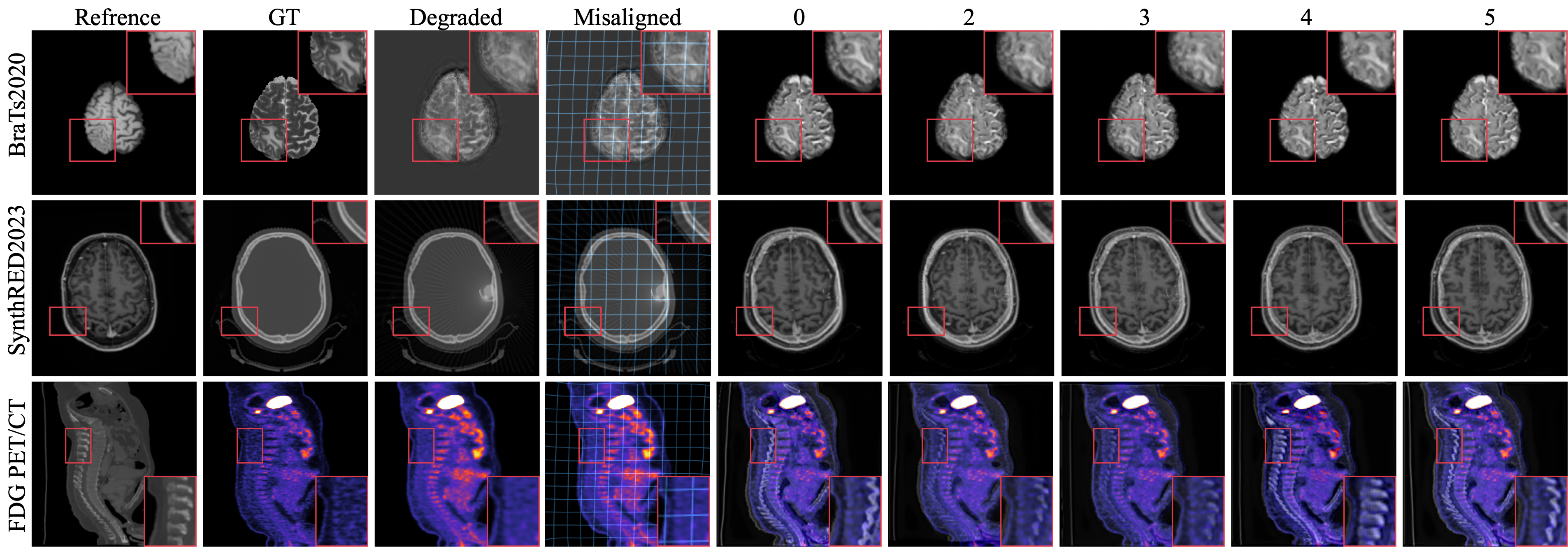}
	\caption{Visual comparison of FA's effectiveness.	The first column shows the reference image without degradation, the second column presents the ground truth (GT) of the degraded image, the third column displays the misaligned degraded image, the fourth column highlights the distortions in the misaligned degraded image, and the fifth to ninth columns show the fusion results under different experimental settings.}\label{c4}
\end{figure*}

\textbf{Effectiveness of OUFR}. To visually validate the effectiveness of OUFR, we replace the Spatial Mamba component with a Transformer and the standard Mamba, respectively, and also conduct an experiment where the entire OUFR is removed for comparison. As shown in Figure \ref{c3}, when Spatial Mamba is replaced by either the Transformer or the standard Mamba, feature misalignment occurs in the fusion results. This phenomenon becomes even more pronounced when the entire OUFR is removed. This is because the introduction of Spatial Mamba effectively eliminates modality differences between source image features, allowing the feature alignment process to proceed without being affected by these differences, thus producing higher-quality fused images. Only when OUFR is fully present does the network achieve optimal fusion quality, further demonstrating the effectiveness of OUFR.

\textbf{Effectiveness of FA}. To visually analyze the impact of using multiple RegBLKs jointly in FA on the fusion results, we conduct an ablation study by adjusting the number of RegBLKs and observing their effects. As shown in Figure \ref{c4}, when no RegBLKs are used in FA, significant feature misalignment occurs in the fused image. When the number of RegBLKs (J) is set to 4, feature alignment achieves optimal results.
\begin{figure*}[ht!]
	\centering
	\includegraphics[height=2.90in,width=6.6in]{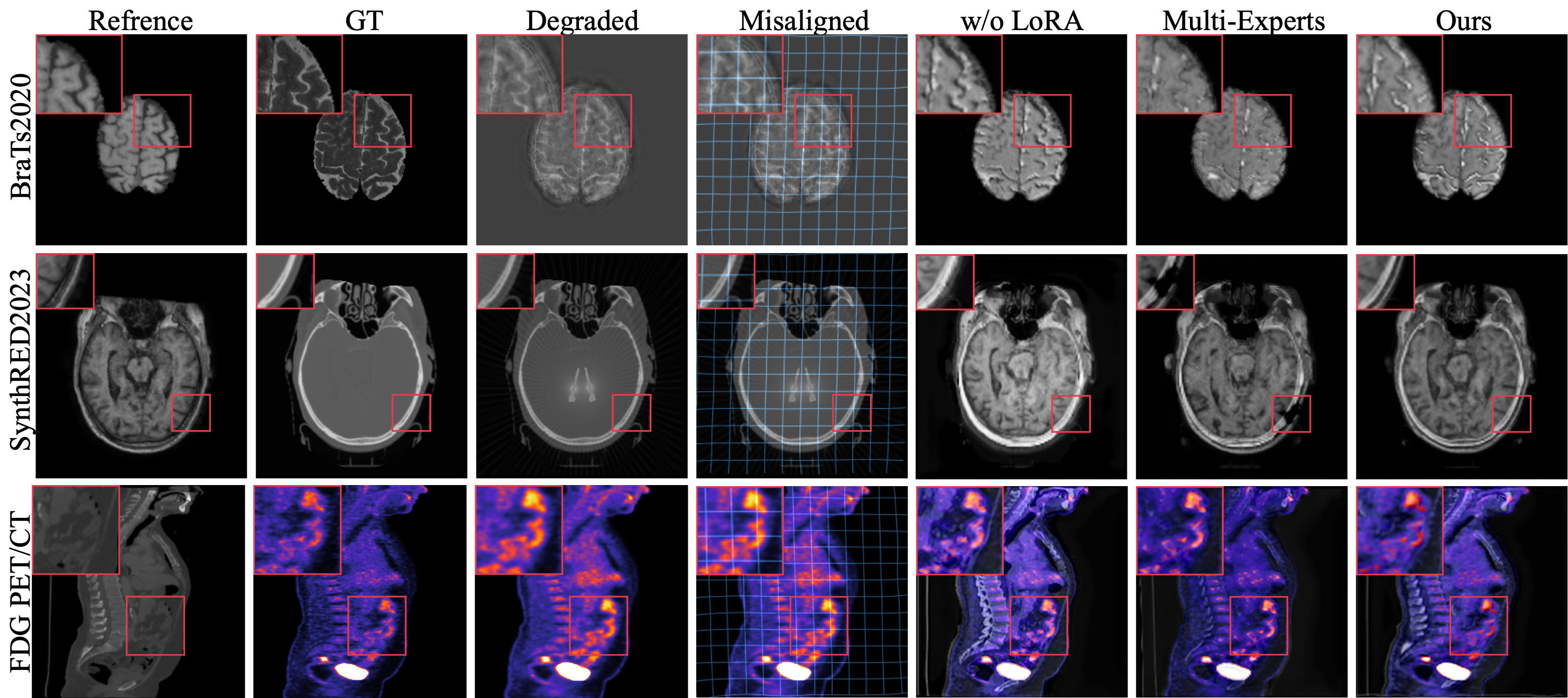}
	\caption{Visual comparison of UFR\&F's effectiveness.	The first column shows the reference image without degradation, the second column presents the ground truth (GT) of the degraded image, the third column displays the misaligned degraded image, the fourth column highlights the distortions in the misaligned degraded image, and the fifth to ninth columns show the fusion results under different experimental settings.}\label{c5}
\end{figure*}

\textbf{Effectiveness of UFR\&F}. To visually demonstrate the effectiveness of UFR\&F, we design two sets of experiments. In the first set, we remove the LoRA branch from ALSN. In the second set, we replace the ALSN branch with a standard multi-expert architecture. As shown in Figure \ref{c5}, after removing the LoRA branch, the network is unable to accurately restore the contrast information in the source images, and the degradation removal effect is compromised due to the loss of the network's adaptive capability to different types of data. When using the multi-expert architecture, feature loss occurs in the fusion results due to the training convergence issues mentioned in the main text. Only when using the complete UFR\&F does the network's fusion results exhibit the best visual quality.

\section{Complexity Comparison}
In practical applications, a model's parameter size and computational load directly determine its deployment difficulty and application cost. Therefore, we conducted a complexity analysis of our proposed method and compared it with existing approaches. Figure 8 presents the complexity comparison results of different methods. Our method achieves optimal performance while exhibiting significantly lower computational complexity than other methods, and its parameter size is also smaller than that of all the compared All-in-One restoration fusion frameworks. This advantage is attributed to our adoption of a single-stage design pattern, which substantially reduces both the number of model parameters and computational overhead. It is worth noting that although our method has more parameters than some single-degradation restoration fusion frameworks, this is due to the incorporation of 3D convolution in the network and the combination of multiple RegBLKs in the FA module.

\begin{figure*}[ht!]
	\centering
	\includegraphics[height=1.65in,width=6.5in]{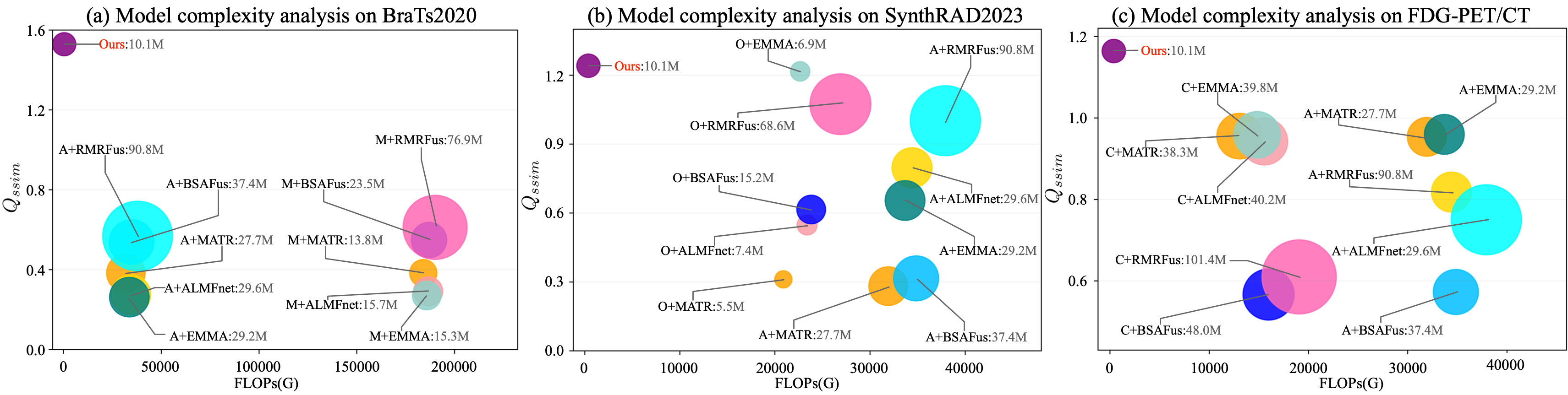}
	\caption{Model complexity analysis. The x-axis represents the FLOPs (in billions) for models processing $256\times 256\times 256$ image inputs, the y-axis denotes the average $Q_{ssim}$ scores, and the bubble size indicates the number of parameters. Subplots (a), (b), and (c) correspond to the experimental results on different datasets.}\label{c6}
\end{figure*}

\section{Limitations of the Method}
Despite the promising performance demonstrated by Unifuse, several limitations remain. First, the method assumes that the input images are degraded but not severely distorted, which may limit its effectiveness when dealing with highly noisy or corrupted images. Second, while Unifuse can handle alignment and fusion in a unified framework, its performance may still be sensitive to extreme misalignments or inconsistencies in image resolutions, as it relies on the assumption of relatively consistent input conditions. Third, the degradation-aware prompt learning module, while effective for a range of common degradation types, may not generalize well to all possible degradation scenarios or unseen image modalities. Finally, while the integration of ALSN allows for adaptive feature representation, its performance could be constrained by the complexity of the network, potentially leading to increased computational cost in resource-limited environments. Further research is needed to address these challenges, enhance the model's robustness, and improve its scalability across diverse real-world applications.
{
    \small
    \bibliographystyle{ieeenat_fullname}
    \bibliography{main}
}

\end{document}